\documentclass[journal]{IEEEtran}

\usepackage{cite}

\usepackage[dvips]{graphicx}

\usepackage{tabularx}
\usepackage{array,multirow}
\usepackage[cmex10]{amsmath}
\usepackage{stfloats}
\usepackage{url}

\mathchardef\mhyphen="2D

\begin{document}

\title{CloudCast: A Satellite-Based Dataset and Baseline for Forecasting Clouds}

\author{Andreas~Holm~Nielsen,
        Alexandros~Iosifidis,
        \IEEEmembership{Senior Member, IEEE}, and~Henrik~Karstoft
\thanks{© 2021 IEEE.  Personal use of this material is permitted.  Permission from IEEE must be obtained for all other uses, in any current or future media, including reprinting/republishing this material for advertising or promotional purposes, creating new collective works, for resale or redistribution to servers or lists, or reuse of any copyrighted component of this work in other works. }%
\thanks{A. Nielsen, A. Iosifidis, and H. Karstoft are with the Department of
Engineering, Aarhus University, Denmark. (e-mail: ahn@eng.au.dk;
ai@eng.au.dk; hka@eng.au.dk)}
\thanks{Digital Object Identifier 10.1109/JSTARS.2021.3062936}}

\ifCLASSOPTIONpeerreview
    \markboth{Journal of \LaTeX\ Class Files,~Vol.~13, No.~9, September~2014}
    {Shell \MakeLowercase{\textit{et al.}}: Bare Demo of IEEEtran.cls for Journals}
\fi

\maketitle

\begin{abstract}
 Forecasting the formation and development of clouds is a central element of modern weather forecasting systems. Incorrect clouds forecasts can lead to major uncertainty in the overall accuracy of weather forecasts due to their intrinsic role in the Earth's climate system. Few studies have tackled this challenging problem from a machine learning point-of-view due to a shortage of high-resolution datasets with many historical observations globally. In this paper, we present a novel satellite-based dataset called ``CloudCast''. It consists of 70,080 images with 10 different cloud types for multiple layers of the atmosphere annotated on a pixel level. The spatial resolution of the dataset is 928 x 1530 pixels (3x3 km per pixel) with 15-min intervals between frames for the period 2017-01-01 to 2018-12-31. All frames are centered and projected over Europe.
 To supplement the dataset, we conduct an evaluation study with current state-of-the-art video prediction methods such as convolutional long short-term memory networks, generative adversarial networks, and optical flow-based extrapolation methods. As the evaluation of video prediction is difficult in practice, we aim for a thorough evaluation in the spatial and temporal domain. Our benchmark models show promising results but with ample room for improvement. This is the first publicly available global-scale dataset with high-resolution cloud types on a high temporal granularity to the authors' best knowledge.
\end{abstract}


\begin{IEEEkeywords}
Remote sensing datasets, spatiotemporal deep learning, atmospheric forecasting.
\end{IEEEkeywords}

%

\section{Introduction}\label{intro}

\IEEEPARstart{C}{loud forecasting} remains one of the major unsolved challenges in meteorology, where cloud errors have wide-reaching impacts on the overall accuracy of weather forecasts \cite{CloudStudy, boucher2013clouds}. Due to the vertical and horizontal nature of clouds, there is an intrinsic difficulty in measuring clouds quantitatively and evaluating the performance of cloud forecasts. This inability to accurately parameterize and thus quantify clouds, convective effects, and aerosols on a sub-grid scale in weather models is one reason model estimates can carry major uncertainties \cite{boucher2013clouds}.

The primary source of quantitative weather forecasts comes from numerical weather prediction (NWP) systems. For these numerical methods, we model the future using governing equations from the field of atmospheric physics \cite{NWP}. Over the past decades, there have been tremendous improvements in weather prediction owing to increased computational power, integration of new theory, and assimilation of large amounts of data. Regardless, these atmospheric simulations are still computationally expensive and operate on coarse spatial scales (9x9 km or above per pixel) \cite{naturedeeplearning, ec}. Furthermore, the current amount of atmospheric data collection exceeds hundreds of petabytes per day \cite{petabyte}, implying that data collection far outpaces our ability to analyze and assimilate it. As a consequence of this, the authors behind \cite{naturedeeplearning} argue that we face two substantial challenges in this field for the future; 1) gaining knowledge from these extreme amounts of data and 2) developing models that tend to be more data-driven compared to traditional approaches while still abiding the laws of physics. One recent application found discrepancies in climate models' estimation of photosynthesis in the tropical rainforests, which ultimately led to a more accurate description of these processes globally \cite{beer2010terrestrial, bonan2011improving}. Ideally, similar insights can be discovered from data-driven methods for cloud dynamics, but obtaining adequate observations of clouds globally has been a substantial obstacle for developing data-driven cloud forecasting methods to-date.

To tackle this problem and spark further research into data-driven atmospheric forecasting, we introduce a novel satellite-based dataset called ``CloudCast'' that facilitates the evaluation of cloud forecasting methods with a global perspective. This approach has been paramount to progress in state-of-the-art methods in the computer vision literature with datasets such as MNIST \cite{mnist}, ImageNet \cite{imagenet}, and CIFAR10 \cite{cifar10}. Current datasets for global cloud forecasting exhibit coarse spatial resolution (9x9km to 31x31km) and low temporal granularity (one-to multiple hours between images) \cite{eumetsat, ec, gfs, cloudimage, cloudanalysis}. We overcome both these issues by using geostationary satellite images, arguably the most consistent and regularly sampled global data source for clouds \cite{CloudStudy}. Since these satellites can obtain images every 5 to 15 minutes with a relatively high spatial resolution (1x1km to 3x3km), they provide an essential ingredient for developing data-driven weather systems, which is an abundance of historical observations. It is possible to achieve higher accuracy in the vertical dimension with radar- and lidar-based profiling methods \cite{cloudsat}, but these falls short on the temporal resolution due to not being geostationary. Our contributions are as follows:
\begin{itemize}
  \item We present a novel satellite-based dataset designed for cloud forecasting.  The dataset has 10 different cloud types for multiple layers of the atmosphere annotated on a pixel level. It consists of 70,080 images with a spatial resolution of 928 x 1530 pixels (3x3 km) and 15-min sampling intervals from 2017-01-01 to 2018-12-31. All frames are centered and projected over Europe. To the authors' best knowledge, no equivalent dataset with high spatial- and temporal resolution exists for evaluating multi-layer cloud forecasting methods globally.
  \item We evaluate four video prediction methods to serve as benchmarks for our dataset by predicting four hours into the future. Two of these are based on recent advancements in machine learning methods specifically for applications in atmospheric forecasting. 
  \item To evaluate our results, we present an evaluation study for measuring cloud forecasting accuracy in satellite-based systems. The evaluation design is based on best practices from the World Meteorological Organization when conducting cloud evaluation studies \cite{CloudStudy}, which includes widely tested statistical metrics for categorical forecasts. Furthermore, we implement the Peak Signal-to-Noise Ratio and Structural Similarity Index from the computer vision literature. The combination of these two domains should provide the best and most fair evaluation of our results.
\end{itemize}

The remainder of the paper is organized as follows. Section \ref{related} provides an overview of the related work. Section \ref{datasetDescription} describes the new dataset. Section \ref{experiments} provides the evaluation study for measuring cloud forecasting accuracy in satellite-based systems. Finally, conclusions are drawn in Section \ref{conclusions}.

\section{Related Work}\label{related}

We start by briefly reviewing related datasets commonly used in the cloud forecasting literature. After presenting related datasets, we will review methods for video prediction that are particularly suitable for forecasting in the spatiotemporal domain.

\subsection{Related Datasets}
We want to introduce a dataset to the community that is particularly suitable for developing data-driven methods with a global perspective. As a result, we only consider 1) geostationary satellite and 2) model-based observations. While other cloud datasets do exist with very high resolution and accuracy for localized atmospheric analysis, such as those from sky-imagers or radar satellites, these technologies generally provide poor spatial coverage, limiting their global use-case.

\subsubsection{Satellite-Based Cloud Observations}
Before introducing related satellite datasets, it is essential to differentiate between cloud detection versus cloud types for meteorological purposes such as forecasting, as the literature varies considerably between the two. Generally speaking, satellite-based cloud detection, typically with the objective of cloud removal, is a relatively established field with many accurate methods \cite{meng2018, meng2020}. For meteorological purposes, most methods typically deal with binary cloud masks \cite{metcloudmask}. As outlined earlier, we are specifically interested in multi-layer cloud types for forecasting purposes, limiting the amount of related literature and datasets quite remarkably. When we focus on multi-layer cloud types, satellite-based cloud observations can be divided into raw infrared brightness temperature and satellite-derived cloud measurements \cite{CloudStudy}. 

Raw satellite brightness temperature acts as a proxy for cloud top height, which will be available during day and night. This proxy is not a perfect indicator for multi-layer cloud types, as the temperature of clouds can also be explained by other factors such as the specific cloud type and seasonal variations. 

Satellite-derived cloud measurements typically involve a brightness-based algorithm that can extrapolate variables such as cloud mask, type, and height from multispectral images \cite{eumetsat}. This is the approach we adopt for our novel dataset, as it can classify multi-layer cloud types with relatively high spatial and temporal resolution in near real-time \cite{eumetsat, nwcsaf}. To the authors' best knowledge, only a few related datasets for satellite-derived multi-layer clouds exist. For the European Meteosat Second Generation (MSG) satellites, these are Cloud Analysis and Cloud Analysis Image published by EUMETSAT \cite{cloudimage, cloudanalysis}. These products exhibit either coarse spatial resolution (9x9km) or infrequent temporal sampling (one to three hours between images) \cite{eumetsat}. As we derive our cloud types directly from the raw satellite images, we can maintain the high resolution (3x3 km) and temporal granularity (15-minute sampling) of the raw satellite images. While our dataset is made from the MSG satellites, our approach is not limited to any specific geostationary satellite system and could be extended for other constellations. Outside Europe, similar datasets exist for a) the American GOES-R satellites called ABI Cloud Height \cite{abicloud} and b) the Japanese Himawari-8 called Cloud Top Height Product \cite{japan}. Due to their geographical coverage not extending to Europe, they are not directly comparable to ours nor those of EUMETSAT.

\subsubsection{Model-Based Cloud Observations}
Model-based clouds are measured using output from an NWP model and is the most directly comparable alternative to satellite observations due to its global coverage. Two commonly used global NWP models are the European ECMWF atmospheric IFS model \cite{ec} and the American GFS model \cite{gfs}. The resolution varies between the two, but the ECMWF model offers the highest spatial resolution with 9x9 km grid spacing \cite{ec}. As both models are global, they can be used interchangeably. The advantage of using NWP model output is that a physics-based simulation of the future exists, while the clear disadvantage is the coarse spatial resolution compared to satellite. Other NWP models also exist on a much finer spatial scale, but these are restricted to local areas, usually on a country-basis \cite{cosmo, arome}.  

Given we want to establish a global reference dataset for data-driven methods, the operational ECMWF model is considered the best global model-based multi-layer cloud dataset. Compared to CloudCast, the ECMWF dataset is inferior in spatial resolution (9x9 km vs. 3x3 km) and temporal resolution (1 hour vs. 15 minutes). Furthermore, the operational ECMWF is not open-source, making it less relevant for the machine learning and computer vision community.

\subsection{Methods}\label{methods}
Producing accurate and realistic video predictions in pixel-space is an open problem to date. Extrapolating frames in the near future can be done relatively accurately, but once the future sequence length grows, so does the inherent uncertainty of the predicted pixel values. Several approaches have been proposed for solving this complex and high-dimensional task: spatiotemporal-transformer networks \cite{sttn}, variational auto-encoders \cite{Denton2018}, generative adversarial networks (GANs) \cite{futuregan, stochasticgan, cloudgan} and recurrent-convolutional neural networks \cite{eid3D, convlstm}. In the video prediction literature, the tasks are often governed by relatively simple physics, such as the Moving MNIST dataset \cite{movingmnist}. However, for predicting atmospheric flow, the task becomes bound by much more complex physics. Therefore, our chosen methods focus on applications that have been explicitly applied for atmospheric forecasting, which will justify the chosen benchmark models for our dataset. These are; 1) \textit{Convolutional Long Short-Term Memory Networks (ConvLSTM)} \cite{convlstm}, 2) \textit{Multi-Stage Dynamic Generative Adversarial Networks (MD-GAN)} \cite{cloudgan} and \textit{$\mathit{TV\mhyphen L^{1}}$ Optical Flow (TVL1)} \cite{urbich2018novel}.

\subsubsection{Convolutional- and Recurrent Neural Networks (ConvLSTM)}\label{convlstm_section}
ConvLSTM was originally developed for precipitation nowcasting using radar images. It is considered the seminal paper for atmospheric forecasting using deep learning, making it relevant as a baseline for our dataset. While newer LSTM-based video prediction methods have been proposed following the ConvLSTM paper, such as PredRNN++ \cite{wang2018predrnn} and Eidetic-3D LSTM  \cite{eid3D}, these were not applied nor evaluated on any atmospheric-related datasets, meaning they are outside the scope defined in Section \ref{intro}. 

\subsubsection{Optical Flow-Based Video Prediction}\label{optflow}
While optical flow is a classical topic in the computer vision literature, it is also one of the most important methods for global data assimilation in meteorology \cite{forsythe2007atmospheric}. Optical flow has been applied for video prediction in several papers \cite{ranzato2014, urbich2018novel}. In \cite{urbich2018novel}, the authors implement the $TV\mhyphen L^{1}$ optical flow method introduced by \cite{zach2007duality} to capture cloud motions on multiple scales spatially based on two subsequent raw satellite images. Once estimated, the flow fields are inverted and applied for extrapolating multiple steps in the future to serve as a forecast of the effective cloud albedo.

\subsubsection{Generative Adversarial Networks}\label{gansection}
GANs have been applied for video prediction in several recent papers \cite{framepred, cloudgan, stochasticgan}. One of these \cite{cloudgan} achieved state-of-the-art results for generating 32-frame time-lapse videos with 128 x 128 resolution of cloud movement in the sky using only one frame as input. The authors use a two-stage generative adversarial network-based approach (MD-GAN), where the first-stage model is responsible for generating an initial video of realistic photos in the future with coarse motion. The second stage model then refines the initially generated video by enforcing motion dynamics using the Gram matrix in the intermediate layers of the discriminator.


\begin{figure*}[!ht]
\centering
\includegraphics[scale=0.73]{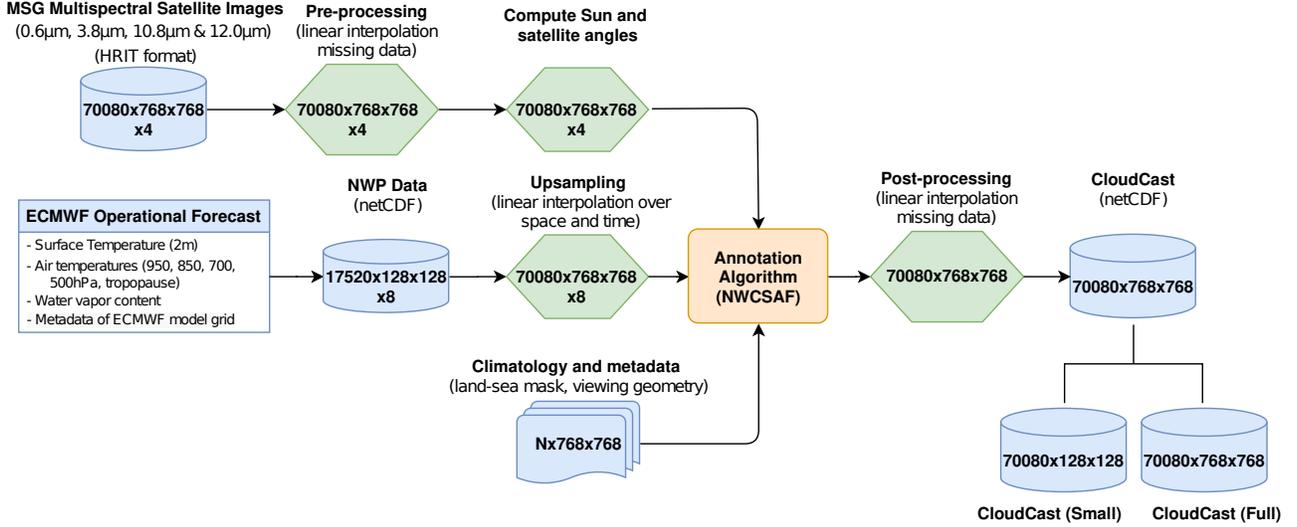} 
\caption{The processing chain going from raw data to our final published dataset, CloudCast. As climatology and land-sea masks are of different resolution, $N$ refers to the specific resolution of each data type, respectively.}
\label{fig:process_chain}
\end{figure*}

\section{Dataset Description}\label{datasetDescription}
The CloudCast dataset contains 70,080 cloud-labeled satellite images with 10 different cloud types corresponding to multiple layers of the atmosphere, as seen in Table \ref{table:1}. As stated in Section \ref{related}, we apply a satellite-derived cloud measurement approach. The procedure for generating our dataset is as follows (see Figure \ref{fig:process_chain} for a visual inspection):

\begin{itemize}
    \item Acquire 70,080 samples for the period 2017-01 to 2018-12. The samples originate from the MSG satellites with four different channels per sample (280,320 satellite images), with 15 minute sampling intervals and  3x3 km spatial resolution. 
    \item Collect hourly NWP output for the entire period using the ECMWF operational model (exact variables to be elaborated shortly).
    \item Annotate the 70,080 samples on a pixel-level using the multi-layer segmentation algorithm (to be described shortly).
    \item Conduct post-processsing to account for short-term missing observations.
    \item Generate and publish a 1) standardized version of the full resolution dataset and a 2) spatially downsampled version in addition to the raw dataset to serve as benchmark for future studies.
\end{itemize}

We will now elaborate on the above steps in more detail. As stated above, we start by collecting the 70,080 raw multispectral satellite images from EUMETSAT. These images come from a satellite constellation in geostationary orbit centered at zero degrees longitude and arrive in 15-minute intervals. The resolution is 3712 x 3712 pixels for the full-disk of the Earth, which implies that every pixel corresponds to a space of dimensions 3x3km. In the remote sensing community, it is well known that infrared channels can observe clouds differently than visible light, meaning infrared is necessary for low- and medium cloud detection. Therefore, we sample one visible channel, two infrared channels and one water vapor channel for each observation to enable multi-layer cloud detection.  The size of the entire raw satellite dataset is around 16 TB. Due to download and request limits imposed by EUMETSAT, we can only process a certain number of samples at any given time, which meant we had to divide this process over a couple of months.

\begin{table}[!t]
\footnotesize   
\centering
\setlength\tabcolsep{3pt}
\caption{Description of the various cloud types present in the CloudCast dataset.}
\begin{tabularx}{\columnwidth}{|>{\hsize=0.5\hsize \centering\arraybackslash}X|
                              >{\hsize=1.5\hsize \centering\arraybackslash}X|}
\hline
\textbf{Class}        & \textbf{Cloud Type}                              \\ \hline
0                     & No clouds or missing data                        \\
1                     & Very low clouds                                  \\ 
2                     & Low clouds                                       \\
3                     & Mid-level clouds                                 \\
4                     & High opaque clouds                               \\
5                     & Very high opaque clouds                          \\
6                     & Fractional clouds                                \\ 
7                     & High semitransparent thin clouds                 \\
8                     & High semitransparent moderately thick clouds     \\
9                     & High semitransparent thick clouds                \\
10                    & High semitransparent above low or medium clouds  \\ \hline
\end{tabularx}

\label{table:1}
\end{table}

Next, we annotate each sample on a pixel level using a segmentation algorithm originally developed by \cite{nwcsaf} under the European Organisation for Meteorological Satellites - Satellite Application Facility on Support to Nowcasting and Very Short Range Forecasting (NWCSAF) project \cite{nwcsafsw}. This algorithm is essentially a threshold algorithm applied at the pixel level for our multispectral satellite images. To improve multi-layer cloud detection in the segmentation algorithm, we include climatological variables and metadata such as geographical land-sea masks and viewing geometry, which have shown to improve low and mid-level cloud detection considerably \cite{nwcsaf}. Additionally, we also include NWP output to further improve the segmentation algorithm by having data not observable from satellite data. We collect the NWP data from the ECMWF operational model, which includes surface temperature, air temperatures at five different heights (950 hPa, 850 hPa, 700 hPa, 500 hPa, and the tropopause level), total water vapor content of the atmosphere, and metadata for the ECMWF model grid. 

Having established all the required datasets for the segmentation algorithm, we will outline how the thresholds are calculated for the major cloud types, which are primarily based on illumination conditions, viewing geometry, geographical location, and the NWP data. We will not list all the specific threshold values due to the sheer number of threshold values, which varies between daytime, nighttime, and twilight. If the reader is interested in these specific values, please see \cite{nwcsaf}.

The first set of clouds is high semitransparent clouds versus opaque (thin) clouds, also called fractional clouds. To separate these, we use the differences between a) the infrared channels $8.7\mu m$ versus $10.8 \mu m$ and b)  channels $10.8\mu m$ versus $12.0 \mu m$. This is based on the insight that these differences are typically larger for cirrus clouds compared to thick clouds \cite{nwcsaf}. To improve separation during daytime where we have visible light available, we apply an additional threshold based on the illumination from the $0.6\mu m$ channel. This can be done due to cirrus clouds having lower reflectance values than opaque clouds with the same radiative temperature. 

Once we have identified semitransparent and fractional clouds, we classify the remaining cloudy pixels into either low-, mid and high clouds found in Table \ref{table:1}. This separation is simpler; hence we can calculate the threshold based on the $10.8 \mu m$ brightness temperature, which directly correlates with cloud height. Atmospheric instability still impacts this separation, which implies we need to correct the previously outlined NWP forecast of air temperatures at different pressure levels. These allow us to separate very low from low clouds, low from medium clouds, medium from high clouds, and high from very high clouds.

We can define the specific thresholds as
\begin{equation}
\begin{split}
vh &= 0.4 * T_{500hPA} + 0.6 * T_{tropo} - 5 K \\
hi &= 0.5 * T_{500hPA} - 0.2 * T_{700hPa} + 178 K \\
me &= 0.8 * T_{850hPA} + 0.2 * T_{700hPa} - 8 K \\
lo &= 1.2 * T_{850hPA} - 0.2 * T_{700hPa} - 5 K
\end{split}
\end{equation}
These thresholds were found in \cite{nwcsaf} to yield the best results. Having defined these thresholds, we can now define our cloud annotation procedure for the remaining cloud types:

\begin{equation}
f(x) = 
\begin{cases}
    VeryHigh & \text{if }  10.8\mu m < vh \\
    High     & \text{if } vh \leq 10.8\mu m < hi \\
    Medium     & \text{if }  h \leq 10.8\mu m < me \\
    Low     & \text{if }  me \leq 10.8\mu m < lo \\
    VeryLow     & \text{if }  lo \leq 10.8\mu m 

\end{cases} 
\end{equation}

In practice the $7.3\mu m$ channel and the secante of the satellite zenith angle is used to reduce the risk of classifying a low cloud as a medium cloud. For notational simplicity we have left these out from the above but they can be found in \cite{nwcsaf}.

While the segmentation algorithm is considered accurate, there are a few limitations. The primary limitation is the scenario where low clouds are sometimes classified as medium clouds in case of e.g., strong thermal inversion, despite the corrections being made with the $7.3\mu m$ channel and solar zenith angle. Nevertheless, we want to highlight that extensive validation of the segmentation algorithm has been carried out using both space-born lidar and ground-based observations to verify its accuracy, which are considered among the most accurate methods for ground-truth observations of clouds \cite{nwcsaf}.

As a final post-processing step, we interpolate missing observations that can arise due to numerous reasons such as scheduled outages or sun outages. More specifically, we interpolate the missing observations from neighboring values linearly, which only happens for short-term periods (below 6 hours). The list of outages at the satellite level can be found at the EUMETSAT website \cite{outages}. One specific example is the 2017-10-17 from 11.30 to 12.30 UTC, where the outage is due to Sun co-linearity.

\begin{figure*}[!t]
\centering
\includegraphics[width=1\textwidth]{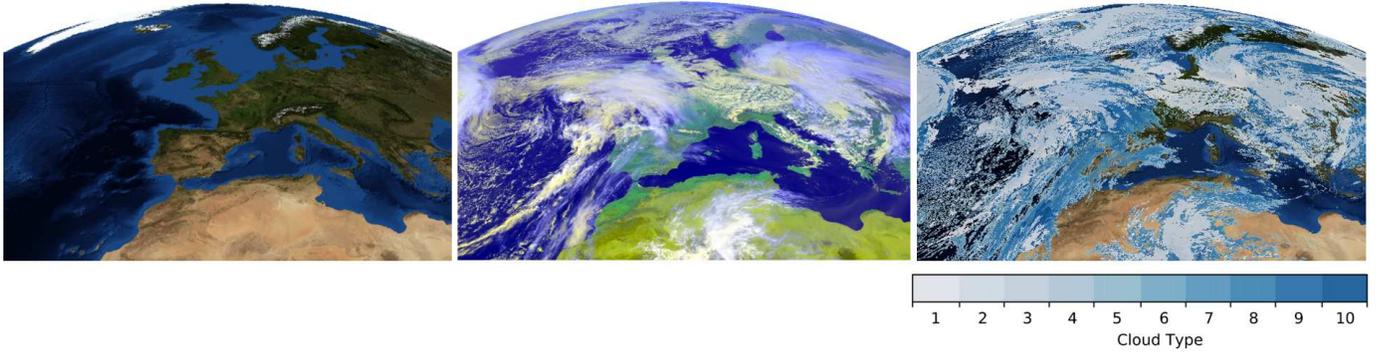}
\caption{Example observation from the CloudCast dataset for the 2017-04-01 13:00 UTC time. \textbf{From left to right:} Raw map of area under investigation; multispectral raw satellite RGB composite consisting of two visible light images (0.6$\mu$m and 0.8$\mu$m) and one infrared (10.8$\mu$m); satellite-labeled CloudCast dataset. The 10 cloud types ("no clouds or missing data" is not assigned a cloud type for visualization purposes) can be seen in the colorbar below the right image.}
\label{fig:dataset}
\end{figure*}

As stated in Section \ref{intro}, current global datasets \cite{gfs, ec} for cloud forecasting and evaluation come with either low temporal granularity (one- to multiple hours between images) or coarse spatial resolution (9x9km to 31x31 km) (see Table \ref{table:comparison}). This demonstrates the need for our novel high-resolution dataset. In addition to the raw dataset, we also publish a standardized version for future studies, where we center and project the final annotated dataset to cover Central Europe, which implies a final resolution of 728 x 728 pixels. An example observation can be seen in Figure \ref{fig:dataset}. To support small-scale experiments and analysis, we also publish a downsampled low-resolution dataset of 15x15 km, which is significantly smaller in size compared to the full dataset. Our novel dataset can be found at \url{https://vision.eng.au.dk/cloudcast-dataset/}.

\newcolumntype{S}{>{\hsize=.23\hsize \centering\arraybackslash}X}
\newcolumntype{M}{>{\hsize=.67\hsize \centering\arraybackslash}X}
\newcolumntype{B}{>{\hsize=.9\hsize \centering\arraybackslash}X}

\begin{table}[t]
\centering
\footnotesize   
\caption{Comparison to existing global-scale datasets for multi-layer cloud types or cloud cover. Note - we have not included the Global Forecast System (GFS) due to its low resolution. Other global NWP models also exist, but none of these compare to the ECMWF model in terms of spatial- and temporal resolution, and therefore excluded from this table.}
\begin{tabularx}{\linewidth}{|M B B S M M M|} \hline
\textbf{Provider}  & \textbf{Dataset Name} & \textbf{Dataset Type} & \textbf{Cloud Types}  &  \textbf{Pixel size} & \textbf{Temporal Frequency} & \textbf{Samples per year} \\ \hline
EUMET\-SAT     & Cloud Analysis MSG  &  Cloud amount, type, phase, height, top temperature  &  3  & 16x16 (km)      & Hourly & 8760 \\ \hline
EUMET\-SAT     & Cloud Analysis Image - MSG  &  Cloud types & 10    & 3x3  & Three hourly   & 2920 \\ \hline
ECMWF     & ERA-5 Reanalysis  &  Multi-layer cloud cover  & 3 & 31x31   & Three hourly  &  2920 \\ \hline
ECMWF     & Operational  &  Multi-layer cloud cover   & 3 & 9x9   & Hourly & 8760  \\ 
\hline
\textbf{Cloud\-Cast (Ours)}     & \textbf{Cloud\-Cast (Ours)}  &  \textbf{Cloud types} & \textbf{10}  &  \textbf{3x3}   & \textbf{15 minutes}  &  \textbf{35040}   \\
\hline \end{tabularx}

\label{table:comparison}
\end{table}

\section{Experiments}\label{experiments}
As an initial baseline study for our CloudCast dataset, we include several of the video prediction methods from our review in Section \ref{related}. These methods have seen considerable success in similar atmospheric nowcasting studies recently \cite{convlstm, urbich2018novel}. 
To match the resolution of most state-of-the-art video prediction methods \cite{fidelity, xingjian2017deep}, we crop and transform our dataset using a stereographic projection to cover Central Europe with a spatial resolution of 128x128. We still use the full temporal resolution of 15-minute intervals compared to hourly observations for other datasets, as mentioned in Section \ref{related}. Several different definitions of nowcasting exist, but they generally vary between 0-2 hours and 0-6 hours \cite{eumetcal, wmosite}. We select the future time frame to be four hours ahead in 15-minute increments (16 time steps), which is somewhere in the middle of most definitions. While forecasting beyond 6 hours is theoretically possible, we expect performance to deteriorate over time unless we incorporate additional variables that cannot be observed from satellite data alone to explain the more medium to long-term cloud dynamics.

We have divided the dataset into 1.5 years (75\%) of training and 0.5 years (25\%) of testing. Ideally, we would want our test data to cover all seasons of the year. However, the frequency distribution between training and test are relatively similar for most classes as seen in Table \ref{table:2}. We also group the 10 cloud types into four based on height: a) no clouds, b) low clouds, c) medium clouds, and d) high clouds. This ensures a more natural ordering of the classes and enables us to focus on the major cloud types also present in the global NWP models \cite{gfs, ec}. 

\subsection{Benchmark Models}
We present an initial benchmark for our dataset based on the reviewed methods in Section \ref{methods}. The results of the baseline models will be presented in Section \ref{results} along with the advantages and disadvantages of the chosen methodologies. 

As stated in Section \ref{methods}, our chosen methods focus on applications that have been explicitly applied for atmospheric forecasting. This is the motivation behind the first three computer vision benchmark models that we outlined in Section \ref{methods}. The final benchmark is the simple persistence model typically used and recommended as a baseline in meteorology studies \cite{CloudStudy}. As these models are all suitable for the problem of cloud forecasting, they should provide good baselines for our dataset.

\subsubsection{Autoencoder ConvLSTM (AE-ConvLSTM)}
For our first baseline, we implement a variant of the ConvLSTM model from \cite{convlstm}, where we introduce an autoencoder architecture with 2D CNNs and use the ConvLSTM layers on the final encoded representation instead of directly on the input frames. This helps us to a) encode the relevant spatial features from the input images before we start encoding and decoding the temporal representation, and b) make training more memory efficient as ConvLSTM layers are memory-intensive. The autoencoder uses skip connections similar to UNet \cite{unet}. The motivation behind including skip connections for video prediction is to transfer static pixels from the input to the output images, making the model focus on learning the movement of dynamic pixels instead \cite{fidelity}.

We start by reconstructing the first 16 input frames to initialize a spatiotemporal representation of the past cloud movement time-series. To predict 16 frames into the future, we use an autoregressive approach, where we feed the predicted output as input recursively to predict the next 16 steps. This is similar to the approach of other video prediction papers \cite{stochasticgan}. To improve the sharpness of our results without introducing an adversarial loss function we have chosen to use the $\ell_{1}$ loss. Furthermore, we use the Adam optimizer with batch size = 4, and we implement an optimization schedule used in the Transformer paper \cite{transformer}, where we increase the learning rate linearly for a number of warmup rounds before decreasing it proportionally to the inverse square root of the current step number. We run the training schedule for 200 epochs with an initial learning rate of $\varepsilon=2e^{-4}$ and momentum parameters $\beta_{1}=0.90$, $\beta_{2}=0.98$ and  400 warmup rounds. We notice only a slight improvement in our loss function after 100 epochs.
\begin{table}[!tbp]
\setlength\tabcolsep{3pt}
\scriptsize   
\caption{Statistical description of the CloudCast dataset. Percentages refer to the total number of observations within the specific column per dataset on a pixel-basis.}
\begin{tabularx}{\columnwidth}{>{\hsize=0.1\hsize \centering\arraybackslash}X
                               >{\hsize=0.5\hsize \centering\arraybackslash}X
                               >{\hsize=0.5\hsize \centering\arraybackslash}X
                               >{\hsize=0.5\hsize \centering\arraybackslash}X
                               >{\hsize=0.5\hsize \centering\arraybackslash}X
                               >{\hsize=0.5\hsize \centering\arraybackslash}X
                               >{\hsize=0.5\hsize \centering\arraybackslash}X}
\hline
&  \textbf{Cloud Type}  & \textbf{Train} & \textbf{Test} & \textbf{Total} & \textbf{Day} & \textbf{Night}  \\ \hline
\multirow{11}{*}{\rotatebox[origin=c]{90}{All}} 
& 0     & 26.5\%         & 32.1\%       & 27.9\%    & 29.5\%   & 26.5\%        \\
& 1     & 12.8\%         & 13.5\%       & 13.0\%    & 13.5\%   & 12.3\%       \\
& 2     & 12.4\%         & 11.2\%       & 12.1\%    & 12.6\%   & 11.7\%     \\
& 3     & 12.0\%         & 9.0\%        & 11.2\%    & 10.7\%   & 11.8\%      \\
& 4     & 9.6\%          & 9.6\%        & 9.6\%     & 9.4\%    & 9.7\%      \\
& 5     & 0.8\%          & 0.7\%        & 0.8\%     & 0.8\%    & 0.8\%       \\
& 6     & 7.8\%          & 7.9\%        & 7.8\%     & 8.4\%    & 7.1\%    \\
& 7     & 3.5\%          & 3.5\%        & 3.5\%     & 3.7\%    & 3.2\%     \\
& 8     & 6.4\%          & 6.2\%        & 6.4\%     & 3.4\%    & 9.6\%     \\
& 9     & 5.6\%          & 4.0\%        & 5.2\%     & 3.8\%    & 6.8\%     \\
& 10    & 2.6\%          & 2.3\%        & 2.5\%     & 4.4\%    & 0.4\%    \\ \hline

\multirow{4}{*}{\rotatebox[origin=c]{90}{Reduced}} 
& No Cloud    & 26.5\%         & 32.1\%        & 27.9\%   & 29.5\%   & 26.5\%      \\
& Low Cloud   & 33.0\%         & 32.6\%        & 32.9\%   & 34.4\%   & 31.1\%     \\
& Mid Cloud   & 12.0\%         & 9.0\%         & 11.3\%   & 10.7\%   & 11.8\%    \\
& High        & 28.5\%         & 26.3\%        & 27.9\%   & 25.4\%   & 30.6\%    \\ \hline

\end{tabularx}

\label{table:2}
\end{table}

\subsubsection{Multi-Stage Dynamic Generative Adversarial Networks (MD-GAN)} \label{mdgan}
For training and optimizing the MD-GAN model, we follow the original authors \cite{cloudgan} with some differences. Since the MD-GAN paper focused on video generation rather than video prediction, we make necessary adjustments to the experimental design to account for this. Instead of cloning one input frame into 16 and feeding them to the generator, we feed the previous 16 images to the generator. 

Besides these changes, we largely followed the approach in \cite{cloudgan}. We found that having the learning rate fixed at $0.0002$ did not produce satisfying results and often caused mode collapse for the generator. Instead, we employed the technique found in the paper \cite{ttur}, where you set a higher learning rate for the discriminator ($0.0004$) than the generator ($0.0001$). This overcomes situations where early mode collapse causes training to stall, and instead incentivizes smaller steps for the generator to fool the discriminator. 
 
We find that the training procedure is inherently unstable, a frequent issue for GANs \cite{improved}. This issue arises particularly for the second stage training, where training seems to stall after around 10-20 epochs.

\subsubsection{$\mathbf{TV\mhyphen L^{1}}$ Optical Flow (TVL1)}
We implement the optical flow algorithm $TV\mhyphen L^{1}$ similar to the authors of \cite{urbich2018novel}, which can capture cloud motion on multiple spatial scales and is one of the most popular optical flow algorithms for meteorological purposes \cite{urbich2018novel}. One of the underlying assumptions of optical flow is constant pixel intensity over time \cite{urbich2018novel}. This is violated due to cloud formation and dissipation. Hence, the presence of convective clouds will negatively impact the accuracy of optical flow algorithms. 
As stated in Section \ref{optflow}, we can extrapolate multiple steps ahead in time using the estimated optical flow recursively on the predicted cloud images. The $TV\mhyphen L^{1}$ algorithm effectively has 11 parameters. The authors \cite{urbich2018novel} optimized these by finding the lowest absolute bias among 21 different parameter settings calculated using a variant of the area-under-the-curve calculation over absolute bias as a function of forecast time \cite{urbich2018novel}. Instead, we conduct a grid search over a hyperspace of 360 different combinations chosen relative to the default values and the optimal hyperparameters found in \cite{urbich2018novel}. As estimated flows (and by extension, the predicted pixel values) will lie in a continuous real-valued space, we round our predictions to the nearest integer in our four-class setting.

\subsubsection{Persistence}\label{persistence}
One of the recommended benchmark models in cloud evaluation studies is called a persistence model \cite{CloudStudy}. Persistence refers to the most recent observation, which in this case is the 15-minute lagged cloud-labeled satellite image, replicated 16 steps into the future. Under the case of limited cloud motion, we expect this model to perform relatively well, but obviously it is naïve and will not work in dynamic weather situations. The most challenging part of video prediction is usually realistic motion generation, and therefore comparing other models to the persistence model shows how well the model has captured and predicted future cloud motion dynamics. Hence, the persistence model will serve as the baseline for skill score calculations.

\subsection{Evaluation Metrics}
Standardized evaluation metrics for the video prediction domain emphasizing atmospheric applications are hard to come by. As stated in Section \ref{intro}, we select our evaluation metrics from the World Meteorological Organization \cite{CloudStudy}. Due to numerous available metrics, we select the ones with the highest ranking score in the referenced paper. As several of these are common in the computer vision and machine learning literature, we only go through the non-standard metrics. Metrics such as Frequency Bias is typically called "bias score", and measures the total number of predicted events relative to observed events. Any value above (below) one indicates the model tends to overforecast (underforecast) events.
\newcommand{\imageforecast}[1]{\includegraphics[width=27mm, height=20mm]{images/forecasts/#1}}
\begin{figure*}[!ht]
\small
\renewcommand{\arraystretch}{2.5}
\setlength\tabcolsep{0pt}
\centering

\begin{tabular}{
    >{\centering\arraybackslash}m{1.2cm}
    >{\centering\arraybackslash}m{2.7cm}
    >{\centering\arraybackslash}m{2.7cm}
    >{\centering\arraybackslash}m{2.7cm}
    >{\centering\arraybackslash}m{2.7cm}
    >{\centering\arraybackslash}m{2.7cm}
    >{\centering\arraybackslash}m{2.7cm}
    }
& \multicolumn{2}{c}{Input Frames} & \multicolumn{4}{c}{Output Frames}     
\\[-1em]
 & $t = -4$   &   \multicolumn{1}{c|}{$t = 0$}      &  $t = 4$   &  $t = 8$  & $t = 12$ & $t = 16$   \\   
\textbf{Ground Truth}   & \imageforecast{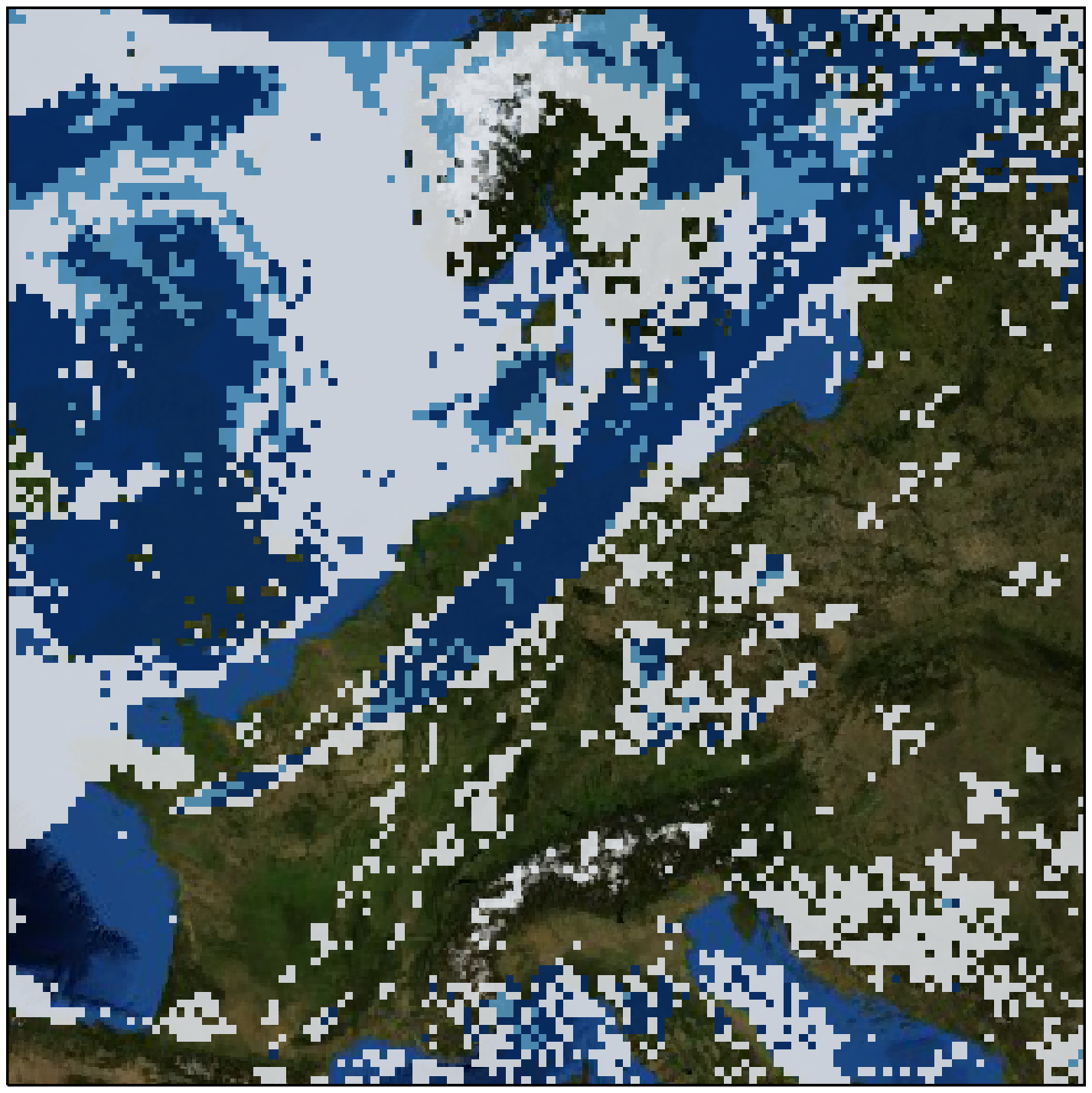}  & \imageforecast{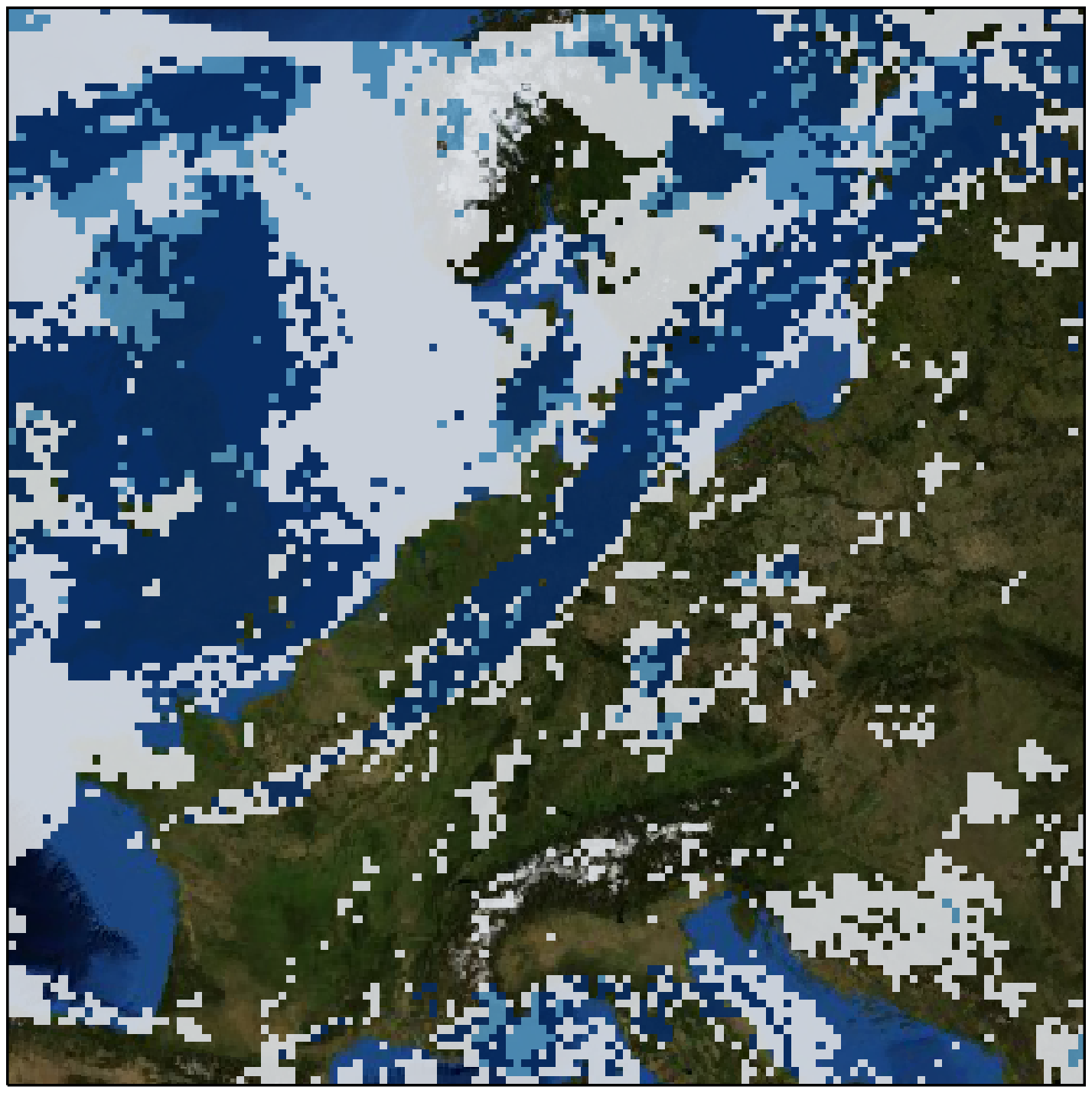}   & \imageforecast{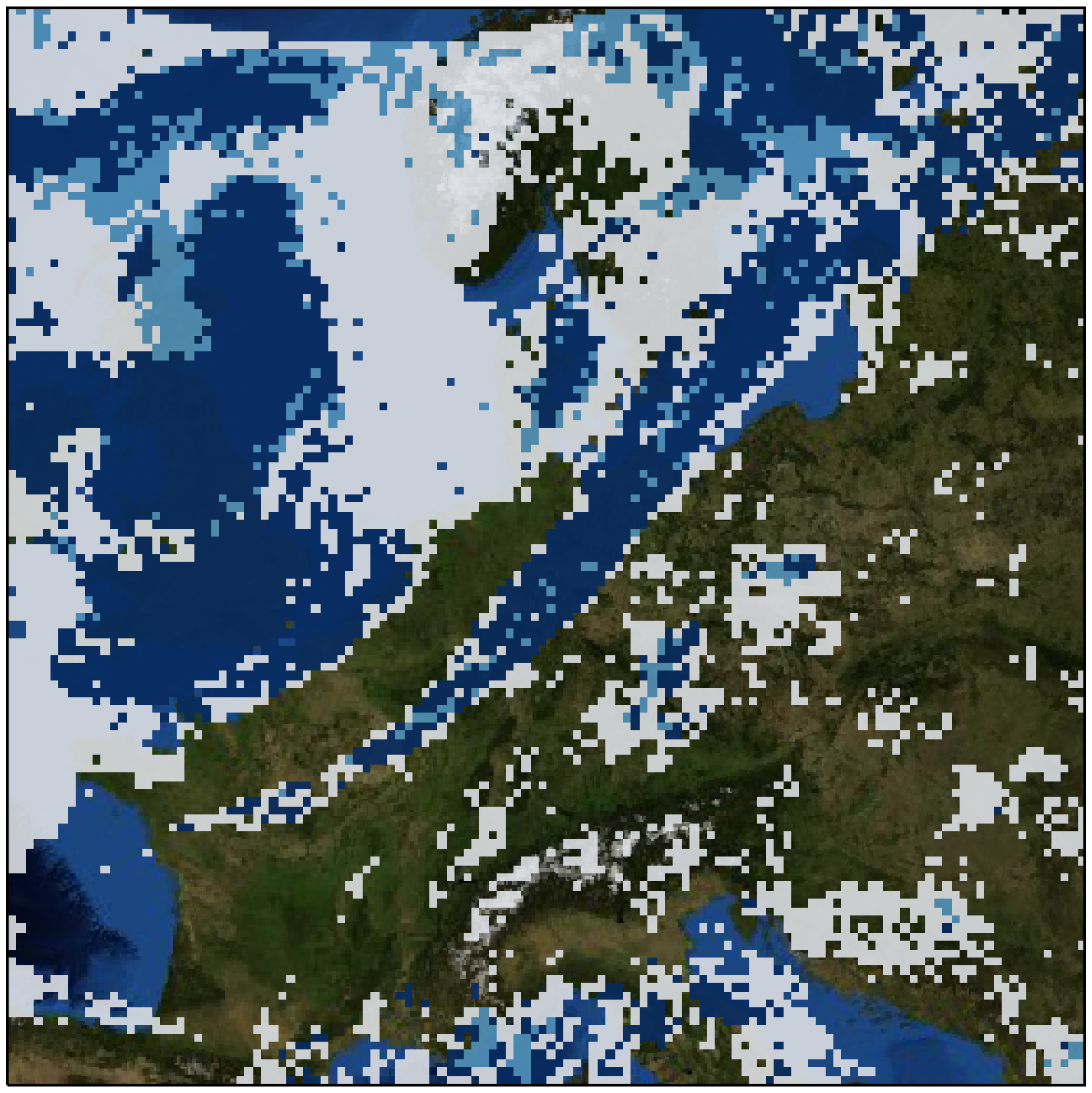}  & \imageforecast{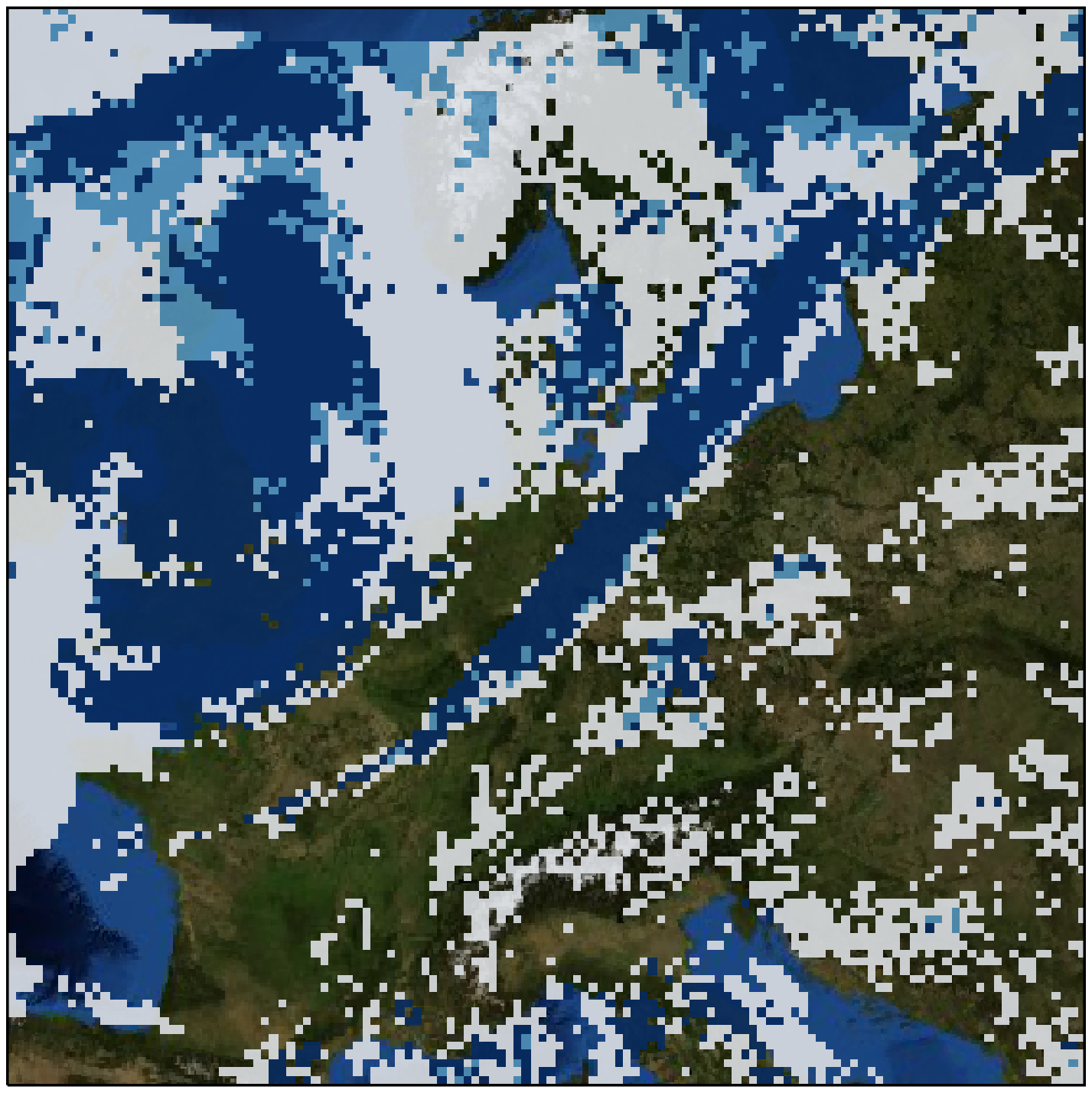}  & \imageforecast{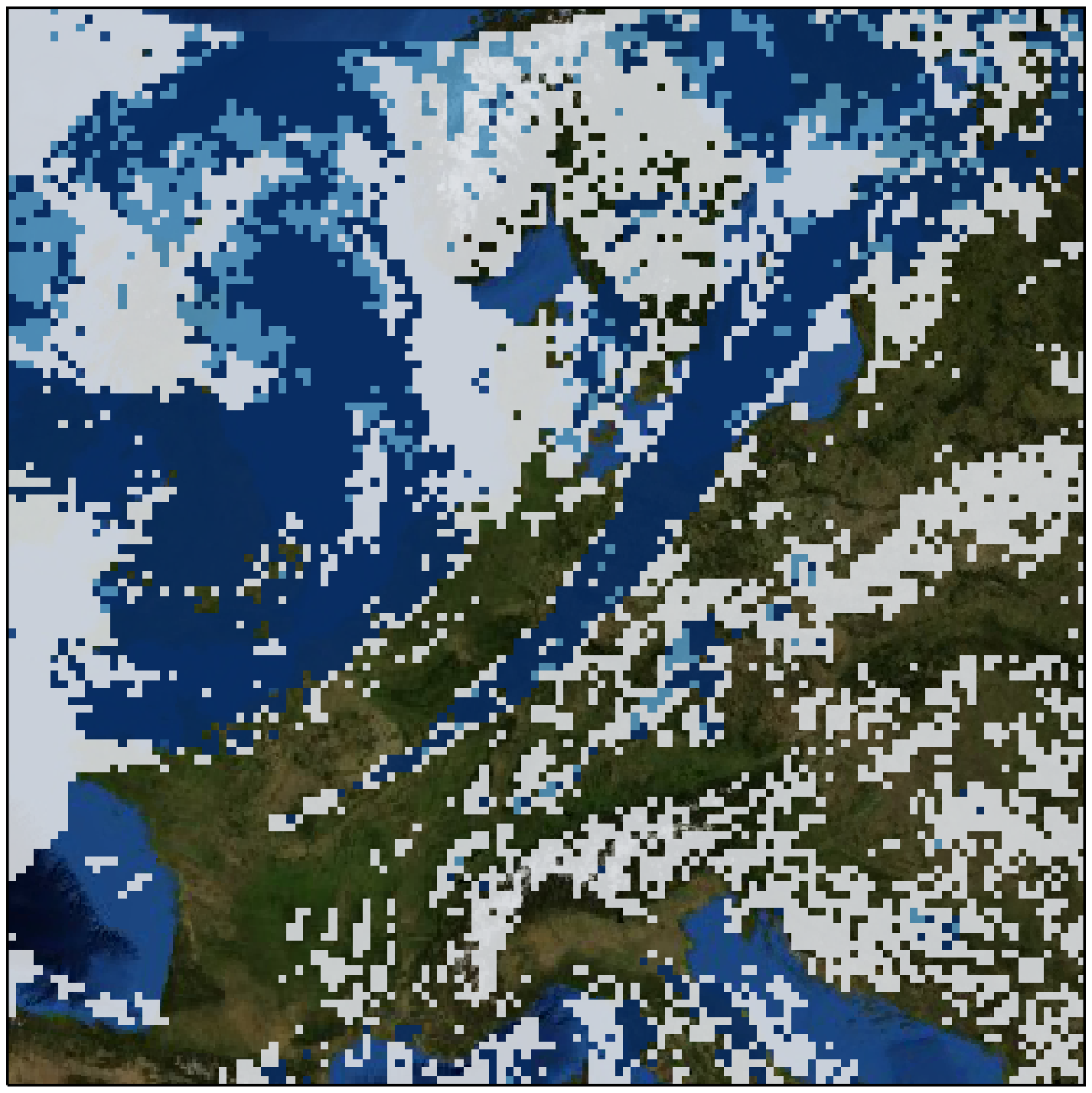}  & \imageforecast{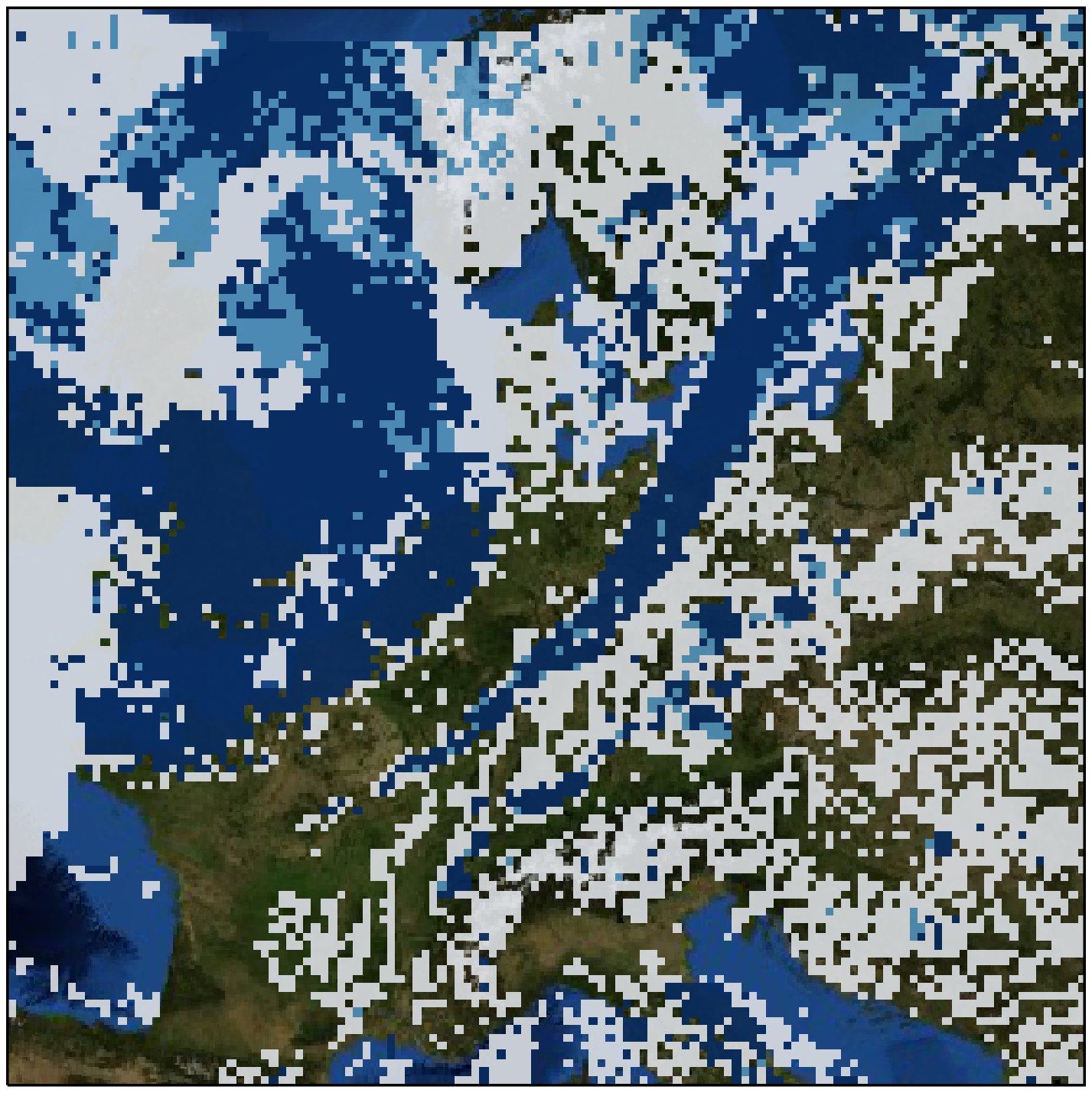}   \\
\\[-3.5em] 
 &   & MD-GAN S2  &  \imageforecast{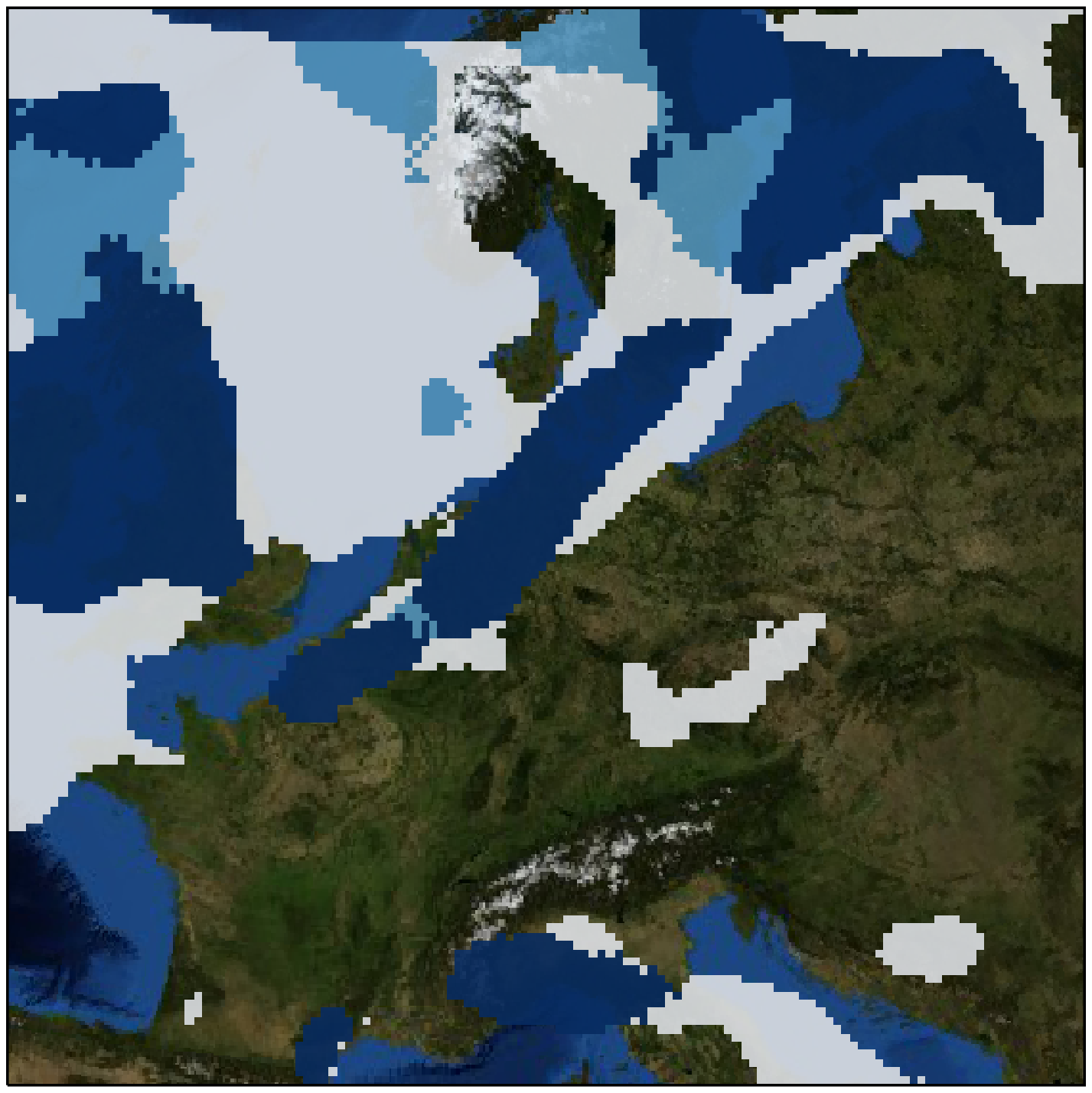}  & \imageforecast{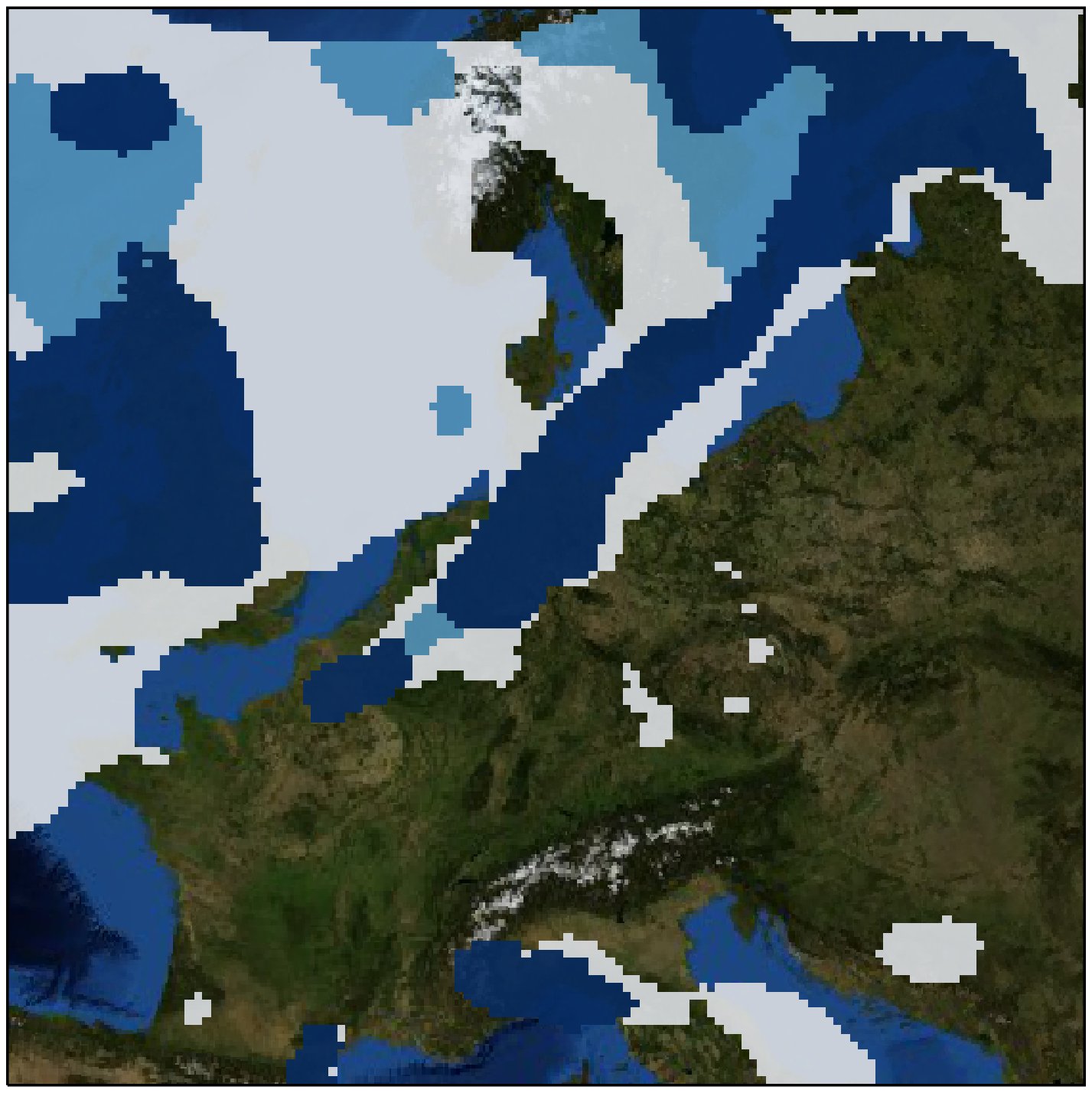}  & \imageforecast{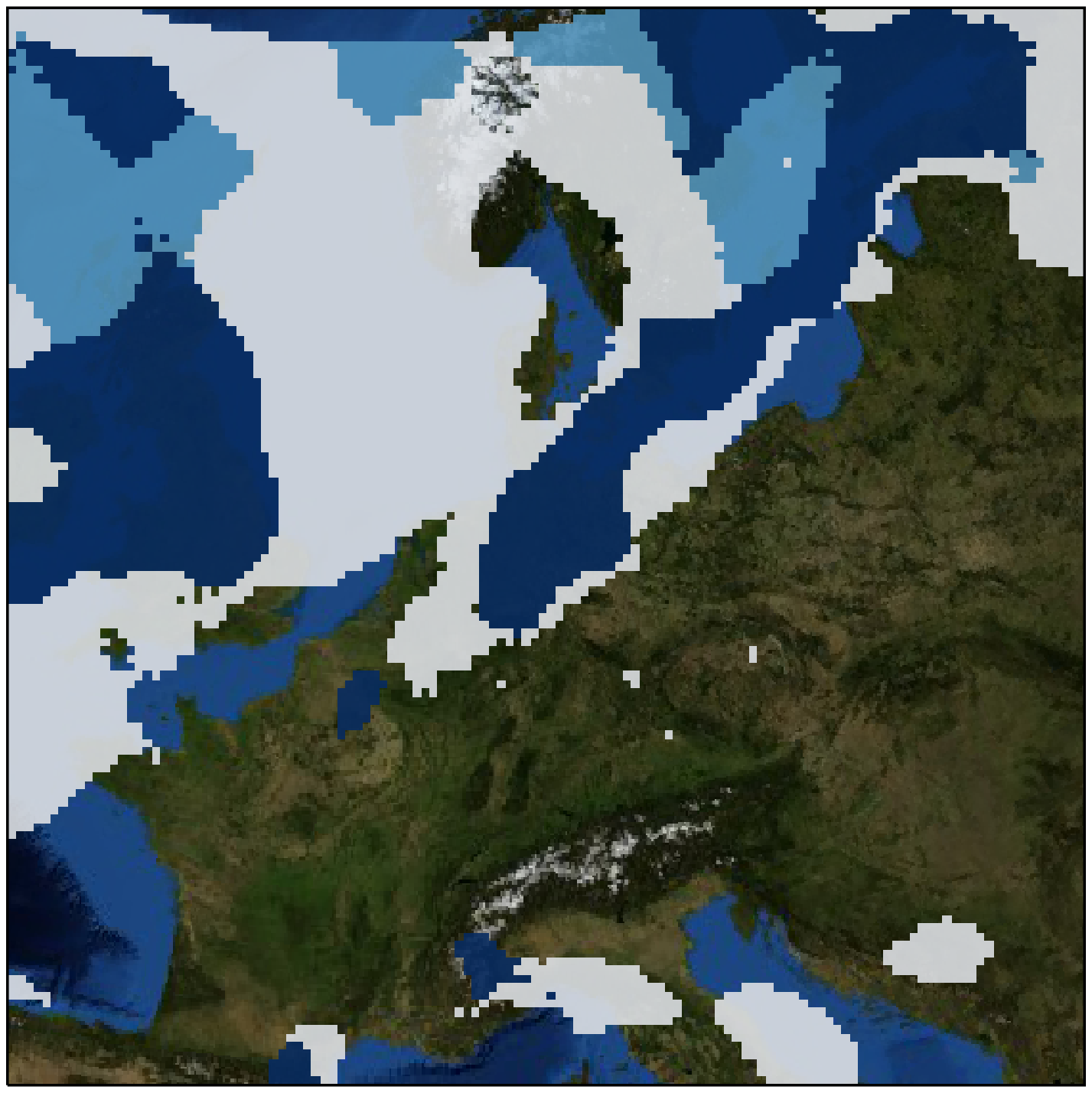}  & \imageforecast{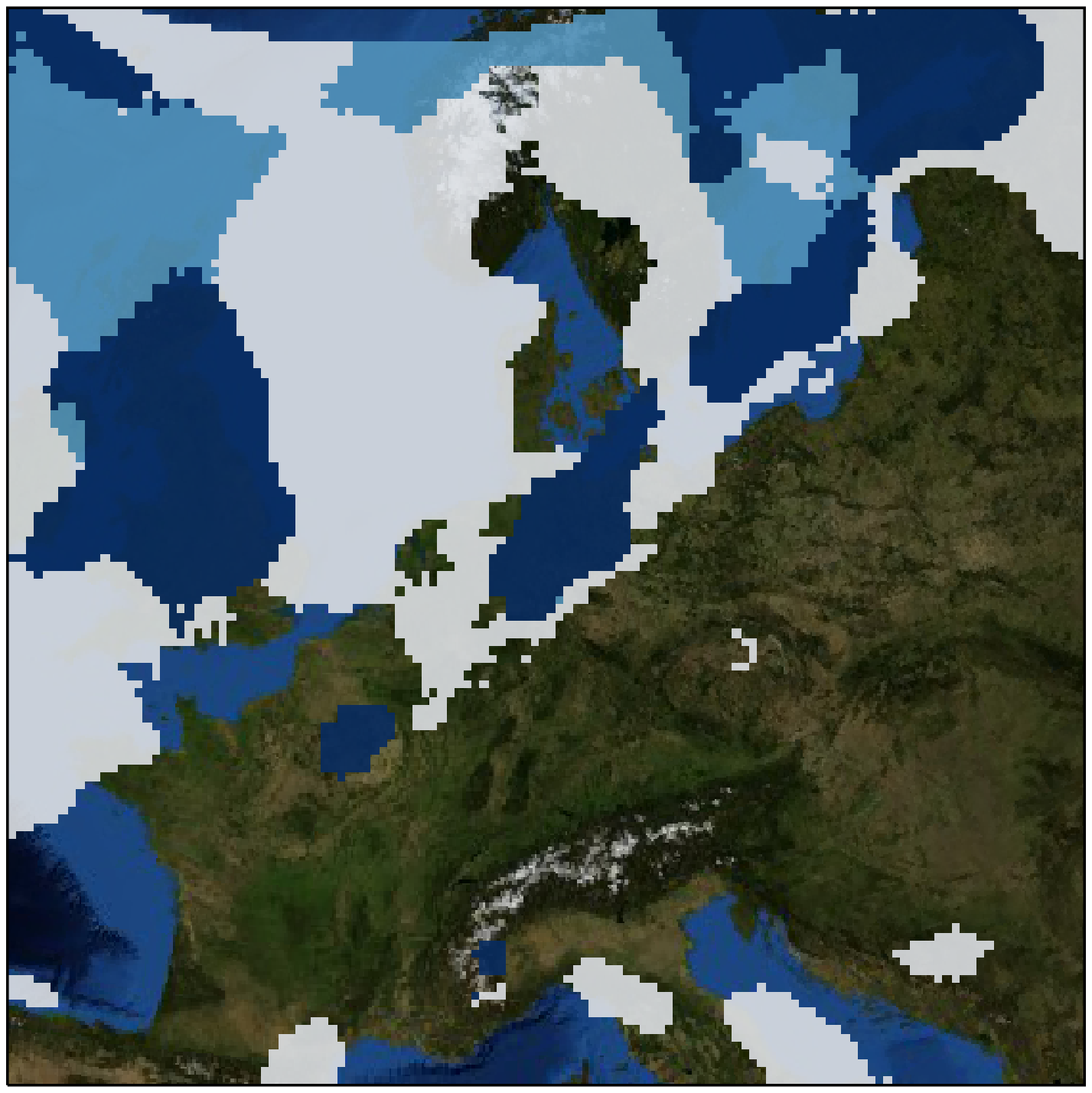}     \\
\\[-3.5em]
  &  & AE-ConvLSTM   & \imageforecast{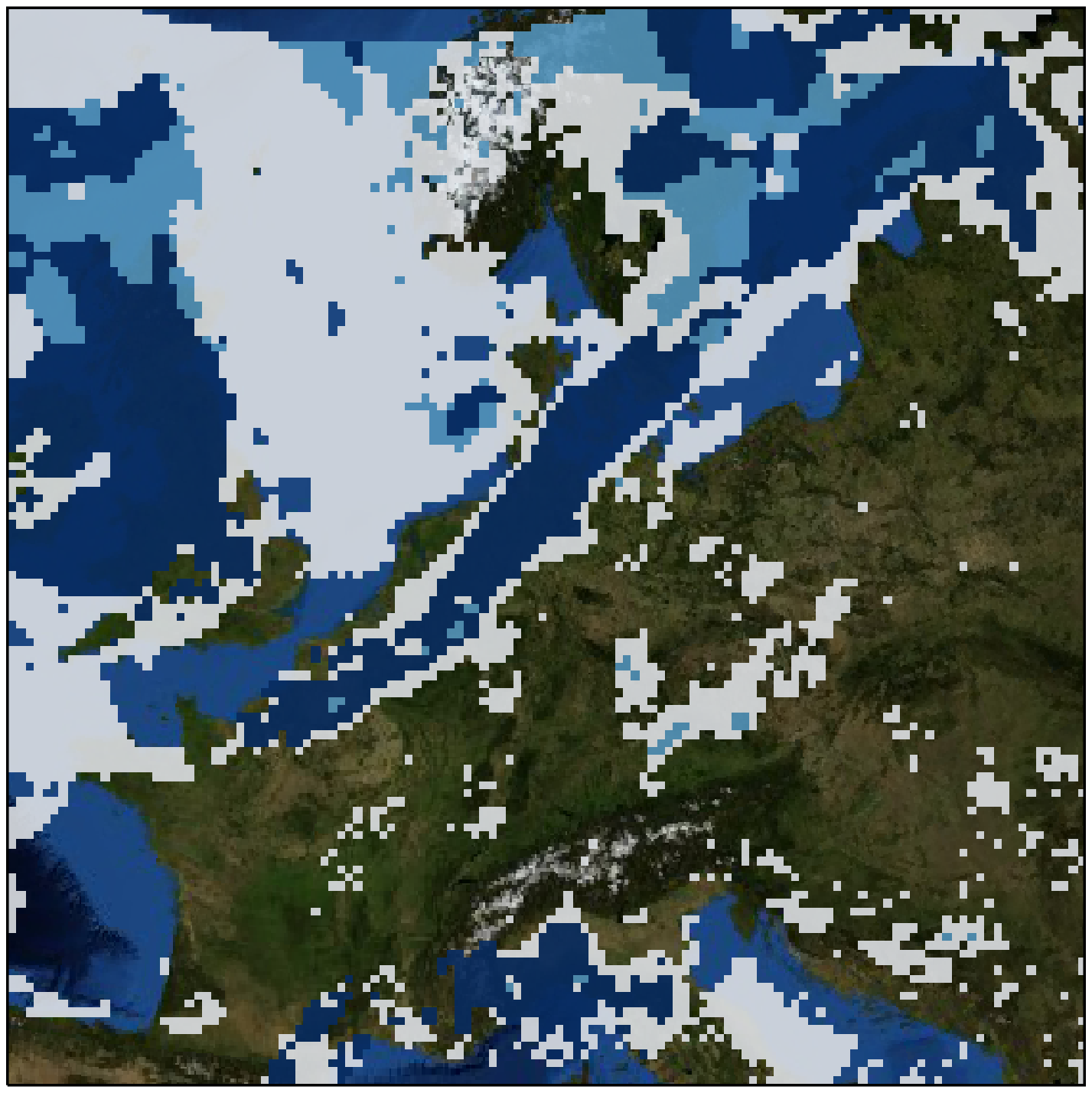}  & \imageforecast{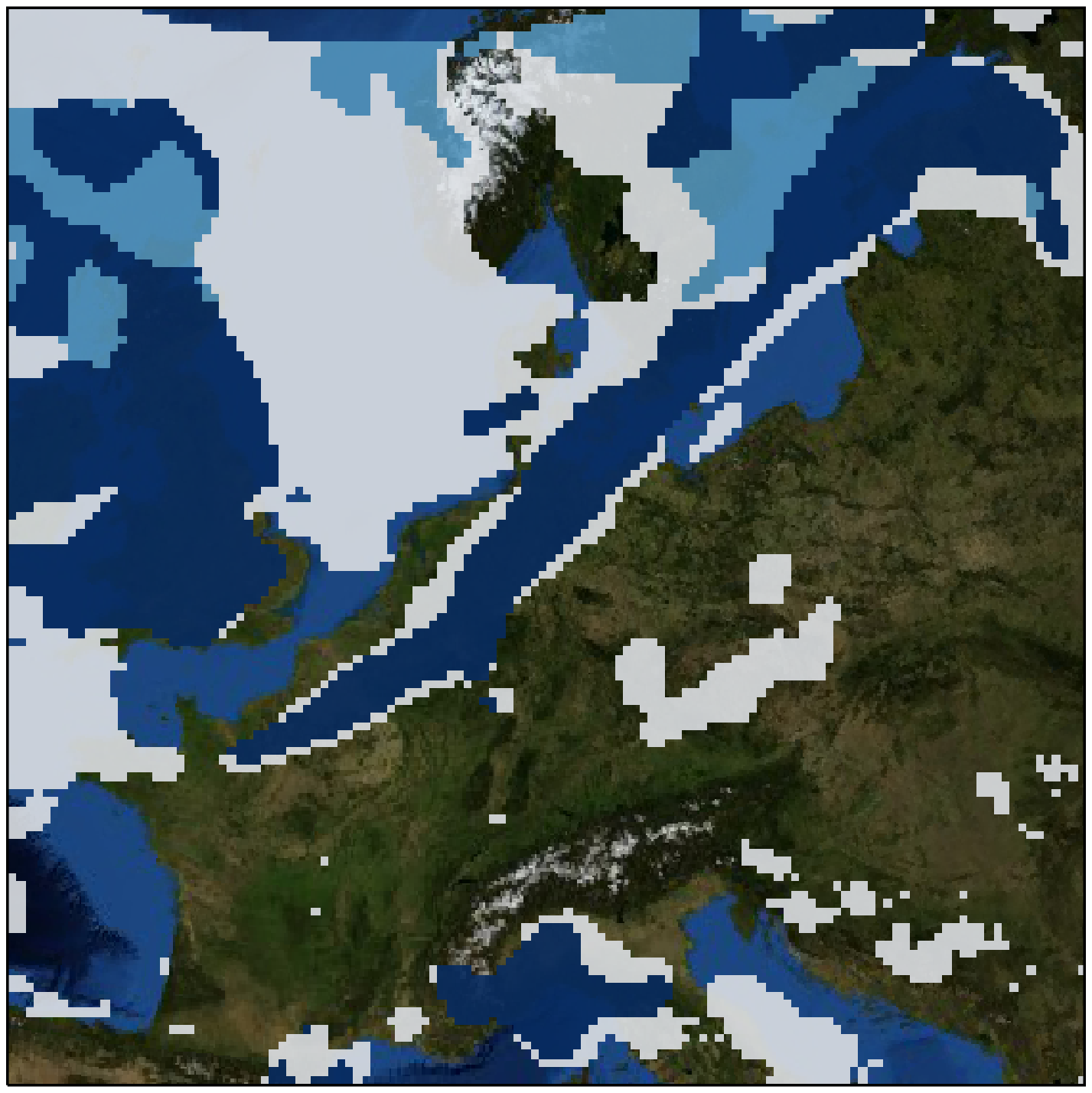}  & \imageforecast{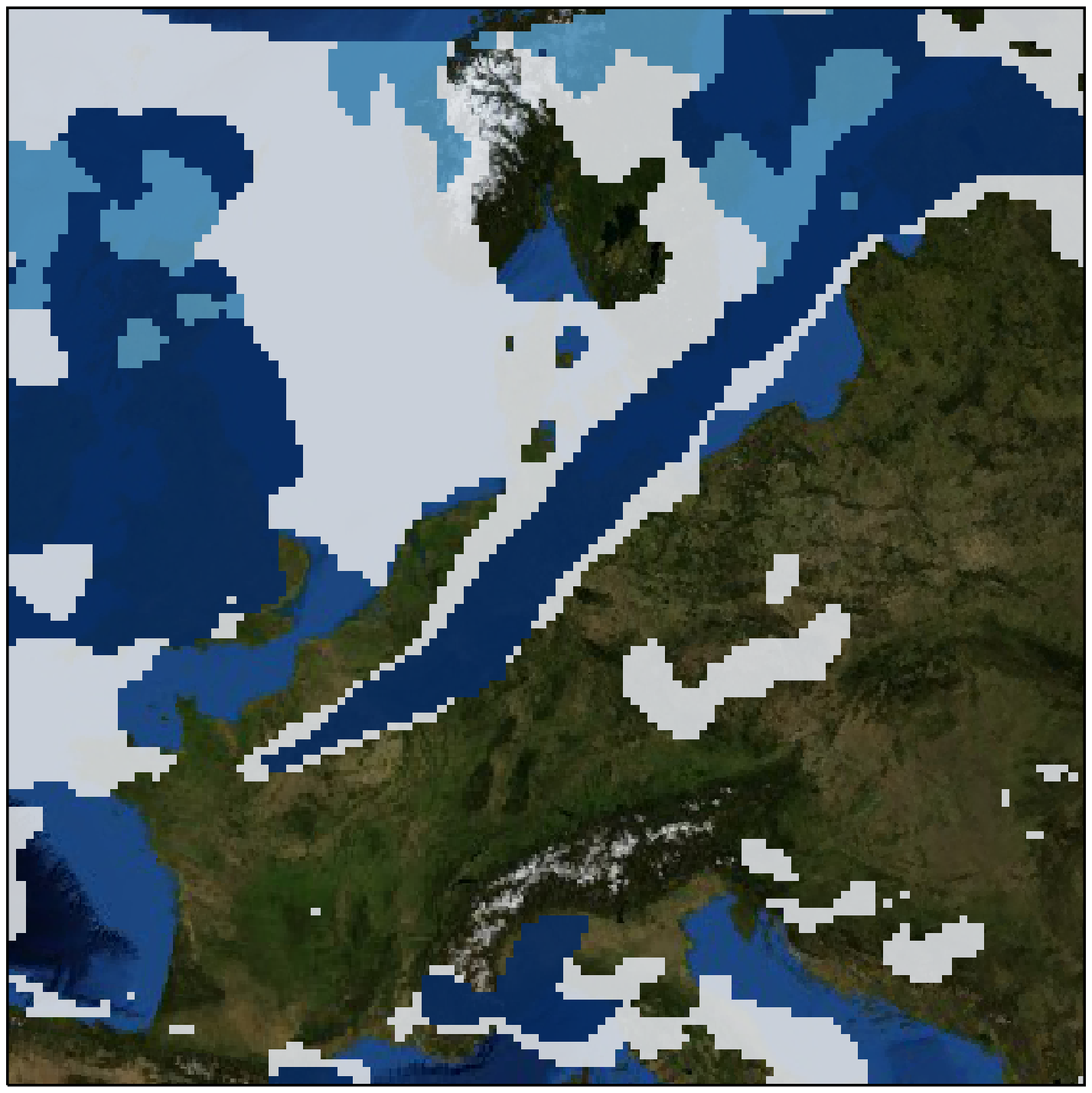}  & \imageforecast{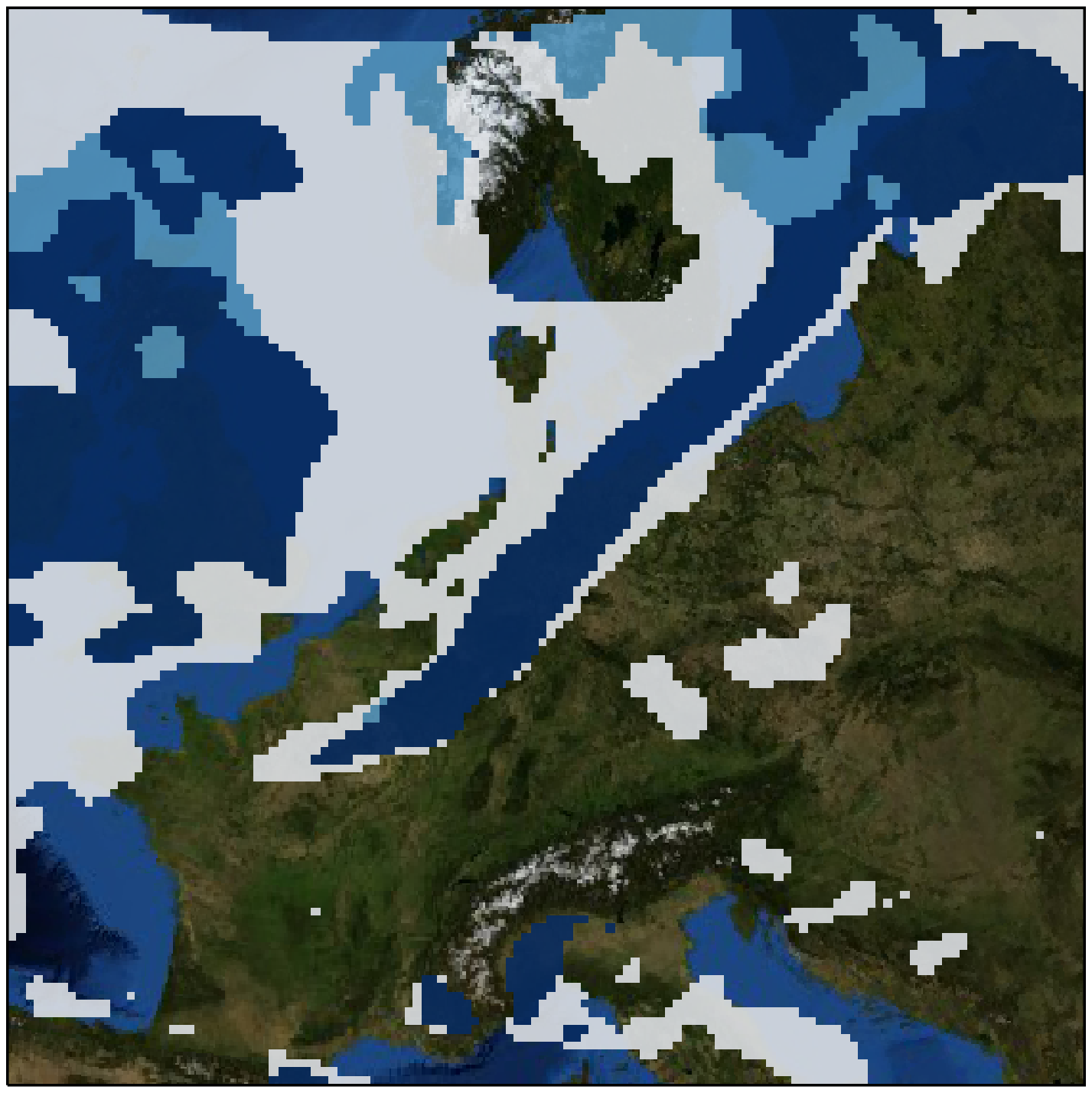}     \\
\\[-3.5em]
  &   &  TVL1  & \imageforecast{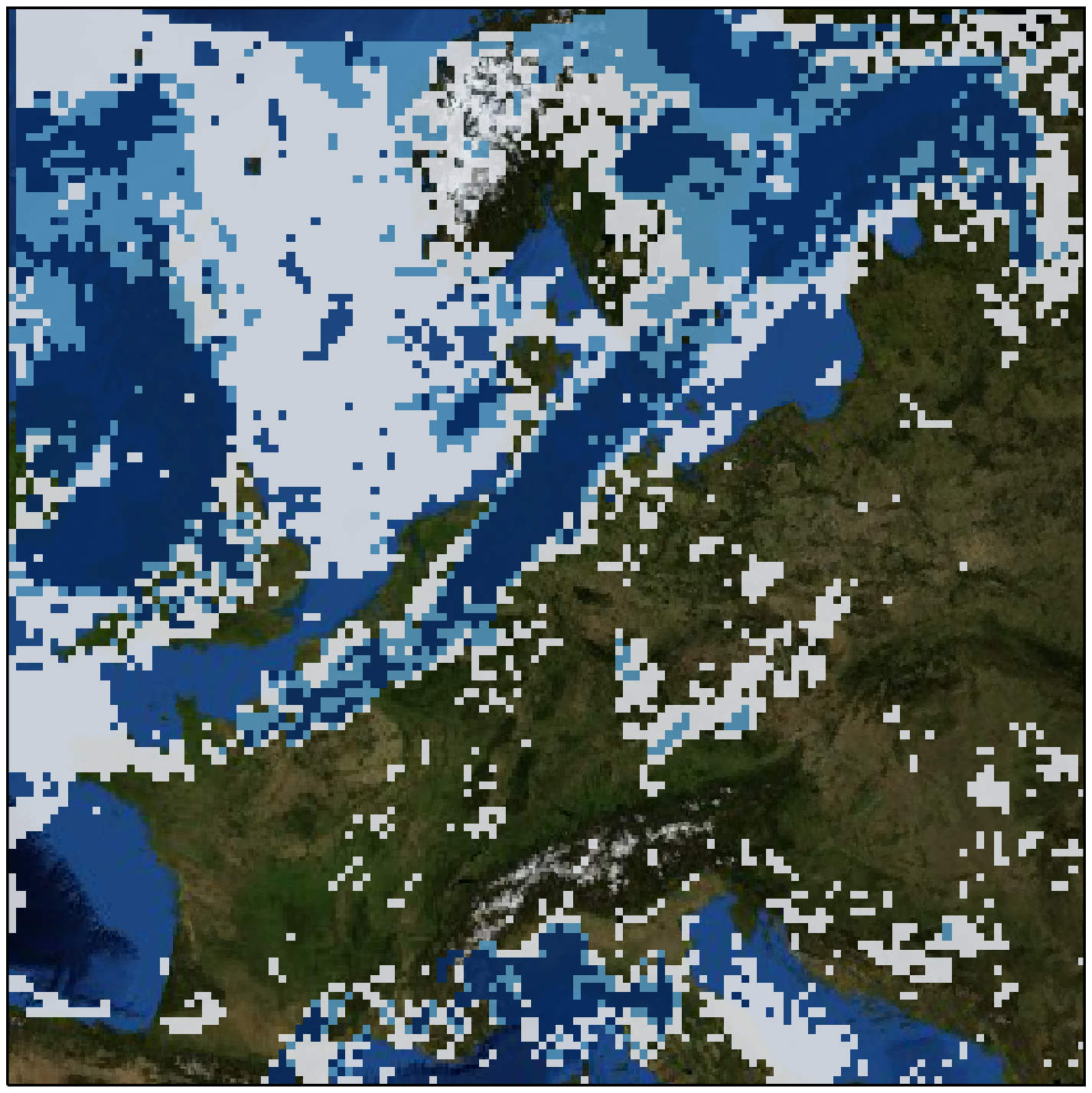}  & \imageforecast{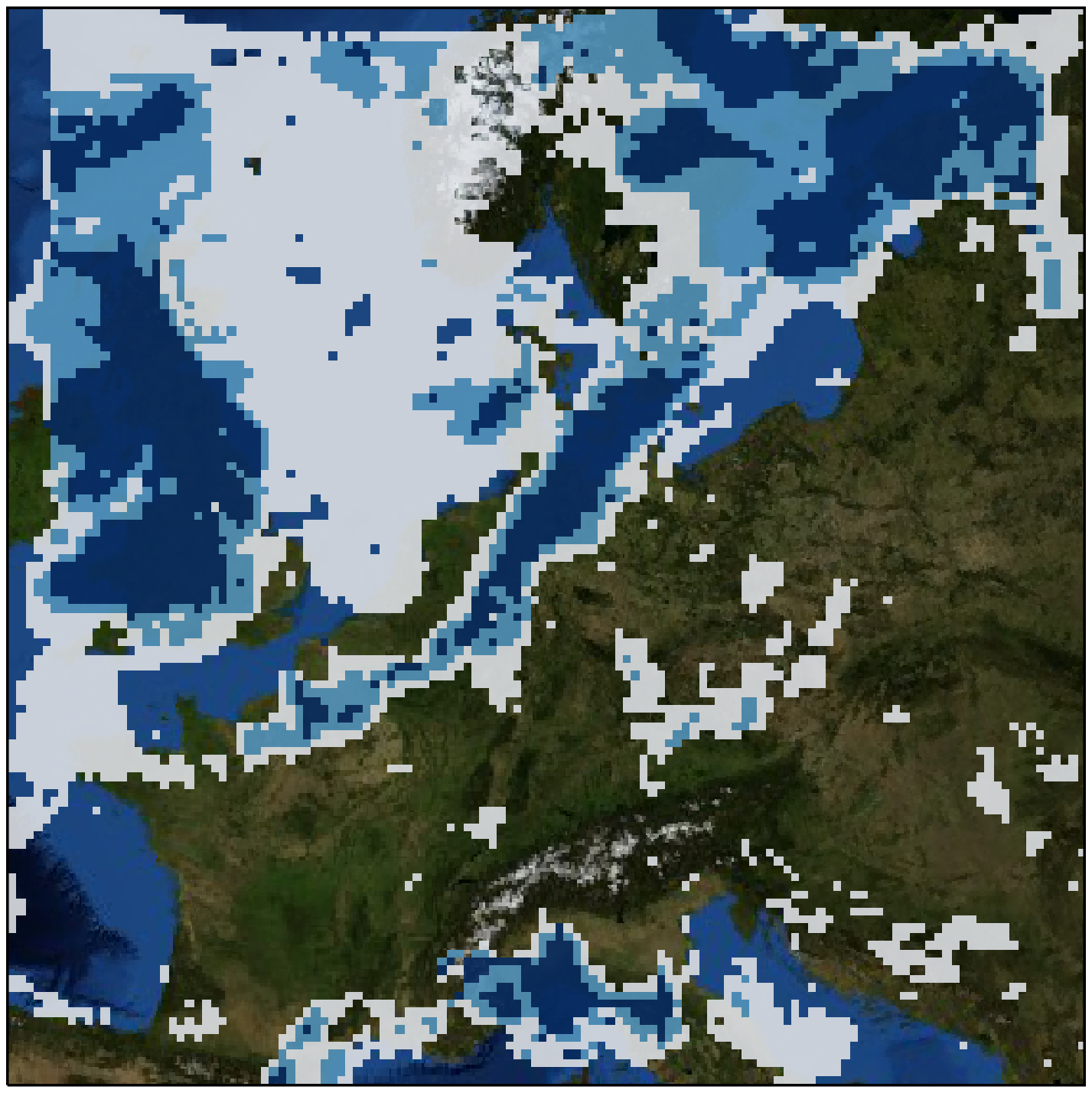}  & \imageforecast{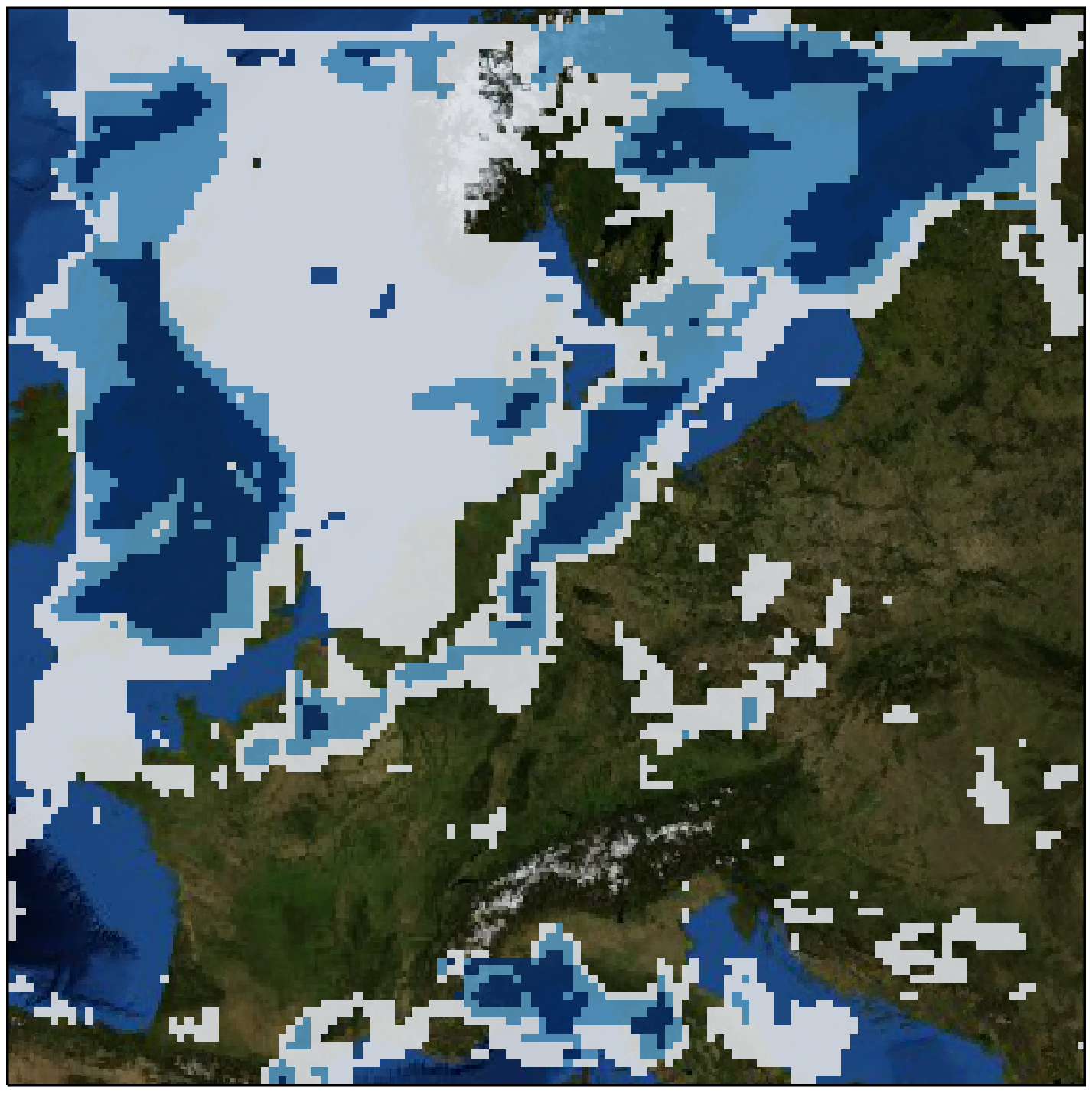} & \imageforecast{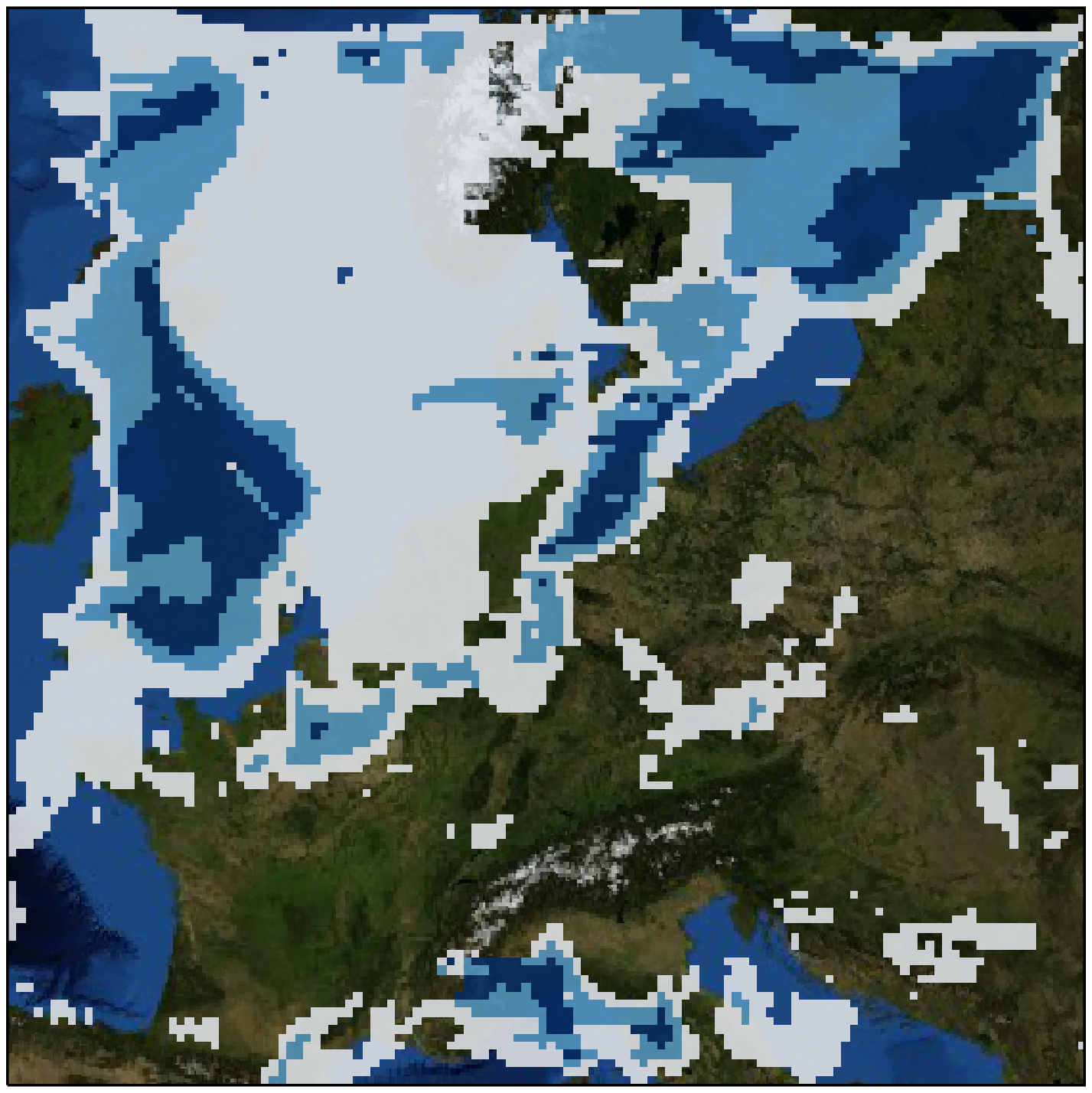}    \\
\\[-3.5em]
    &   & Persistence  & \imageforecast{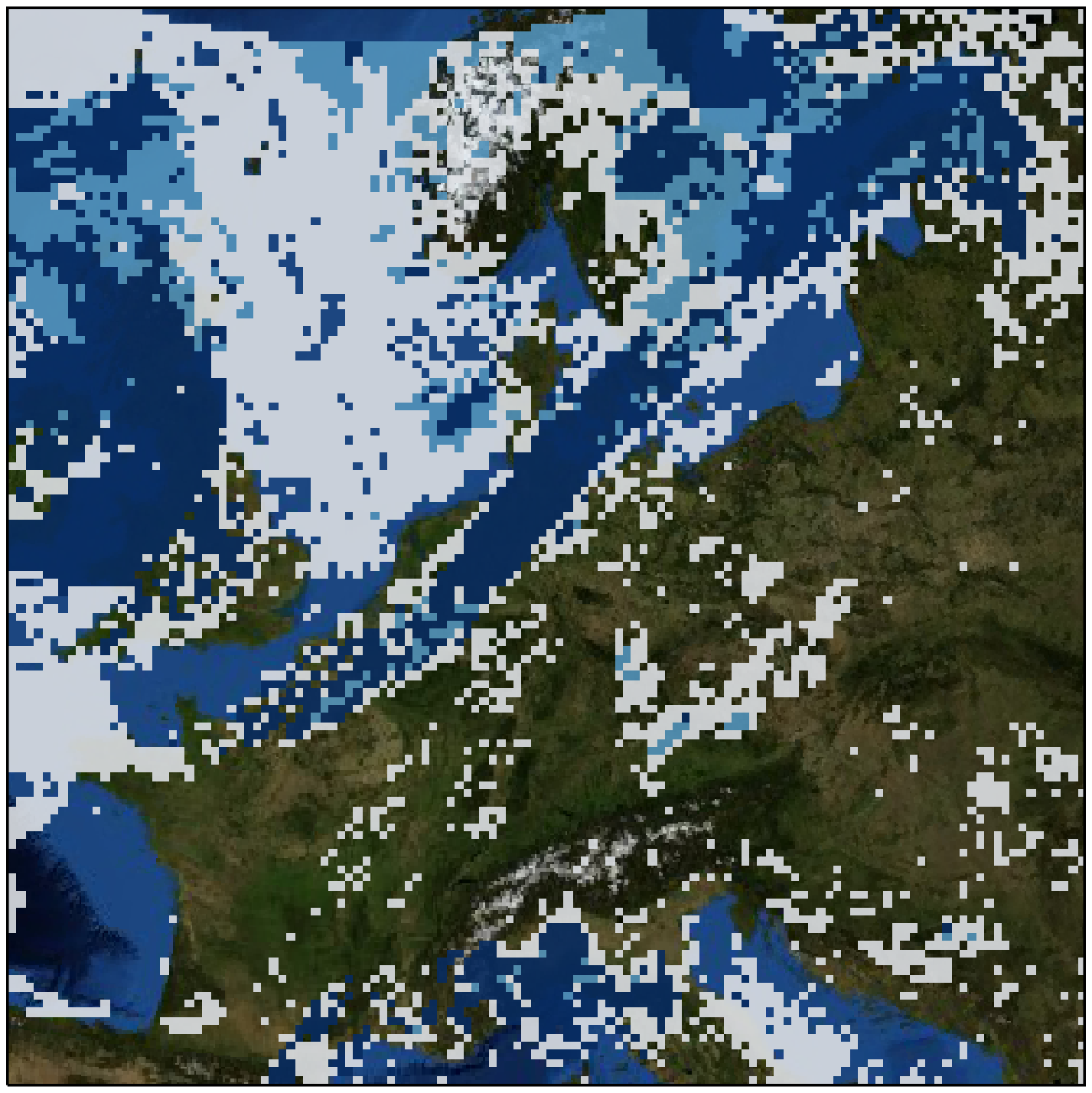}  & \imageforecast{Persistence_frame_0.eps}  & \imageforecast{Persistence_frame_0.eps}  & \imageforecast{Persistence_frame_0.eps}  
\\[-1em]
\end{tabular}

\caption{Example forecast for all models relative to ground truth. $t$ refers to forecast time in 15-minute increments. Here we include two input images corresponding to two- and one-hour before. We also include four output images corresponding to one- to four hours ahead predictions in one-hour increments.}
\label{fig:forecast}
\end{figure*}

The first non-standard metric is called  ``Brier Score''. In this case, the Brier Score refers to the MSE between estimated probabilistic forecasts and binary outcomes. To extend it to the categorical multi-class setting, we sum the individual MSE's for all categorical probabilistic forecasts relative to the one-hot target class variable as follows
\begin{equation}
B S= \frac{1}{M} \frac{1}{N} \sum_{k=1}^{M}  \sum_{t=1}^{N}\left(f_{t,k}-y_{t,k}\right)^{2},
\end{equation}
where $f_{t,k}$ is the predicted probabilities for all class $k$ pixels in a given image at time $t$,  $y_{t,k}$ the actual binary outcomes, $N=16$  the number of future time steps and $M=4$ the number of classes. 
The second is called Brier Skill Score, and it is calculated using the MSE of a given model relative to some benchmark forecast. As mentioned in Section \ref{persistence}, we use the persistence forecast as the benchmark in skill score metrics. The formula for the Brier Skill Score is

\begin{equation}
BSS= 1 -\frac{BS_{model}}{BS_{persistence}},
\end{equation}
where any value above (below) zero implies superior (inferior) performance by the proposed model.

In addition to these metrics, we also include video prediction metrics from the computer vision literature, which, taken together with the meteorology metrics, should constitute the fairest evaluation in this complex setting. These include the Structural Similarity Index (SSIM) and the Peak Signal-to-Noise Ratio (PSNR).

\begin{table}[!t]
\scriptsize   
\renewcommand{\arraystretch}{1.3}
\setlength\tabcolsep{3pt}
\caption{Key evaluation metrics based on \cite{CloudStudy} for our proposed AE-ConvLSTM model, the second-stage MD-GAN model, the $TV\mhyphen L^{1}$ optical flow model and the simple persistence model. Reference forecast used in Brier Skill Score metric is the persistence model. SSIM and PSNR refer to the Structural Similarity Index and Peak Signal-to-Noise Ratio, respectively.}
\begin{tabularx}{\columnwidth}{>{\hsize=0.6\hsize}X
                               >{\hsize=0.5\hsize \centering\arraybackslash}X
                               >{\hsize=0.3\hsize \centering\arraybackslash}X
                               >{\hsize=0.3\hsize \centering\arraybackslash}X
                               >{\hsize=0.3\hsize \centering\arraybackslash}X}
\hline
\textbf{Metric} & \textbf{AE-ConvLSTM}  & \textbf{MD-GAN S2}  & \textbf{TVL1}  & \textbf{Persistence}  \\ \hline
Mean Accuracy             & 68.44\%    & 67.06\%     & 64.30\%  & 63.60 \%   \\
Frequency Bias            & 0.96       & 0.95        & 1.10     & 1.00      \\
Brier Score               & 0.16       & 0.16        & 0.18     & 0.18      \\
Brier Skill Score         & 0.11       & 0.07        & 0.02     & NA      \\ 
SSIM                      & 0.66       & 0.60        & 0.58     & 0.55      \\ 
PSNR                      & 8.06       & 7.83        & 7.50     & 7.41      \\ \hline
\textbf{Spatial Accuracy} &            &             &          &      \\
No Clouds                 & 74.09\%    & 73.88\%     & 67.98\%  & 70.86\%      \\
Low Clouds                & 66.34\%    & 65.31\%     & 64.48\%  & 61.90\%      \\
Medium Clouds             & 39.90\%    & 34.42\%     & 56.70\%  & 45.32\%      \\
High Clouds               & 73.90\%    & 73.28\%     & 63.86\%  & 66.50\%      \\   \hline
\textbf{Temporal Accuracy} &    &             &          &           \\
1-Hour Forecast           & 73.58\%    & 72.32\%     & 70.07\%  & 69.49\%   \\
2-Hour Forecast           & 67.39\%    & 67.36\%     & 63.43\%  & 62.50\%   \\ 
3-Hour Forecast           & 63.44\%    & 62.88\%     & 58.89\%  & 57.88\%   \\ 
4-Hour Forecast           & 60.46\%    & 59.28\%     & 55.23\%  & 54.27\%   \\ \hline
\end{tabularx}

\label{result_table}
\end{table}

\subsection{Results}\label{results}
The results of our baseline methods can be found in Table \ref{result_table}. We also include an example of model forecasts relative to ground truth in Figure \ref{fig:forecast}. 

Despite the proposed models showing relatively high overall accuracies, it is quite clear that none of the models show consistent performance across time and space. This is also evident when looking at the decline in accuracy across time. This suggests that we need to a) develop models more suitable for this particular problem or b) incorporate other data sources or variables to make more reasonable and causal predictions for the complex setting of multi-layer cloud movement and formation.

We include a visualization of the worst predictions from the test set measured by mean accuracy in Figure \ref{fig:bestworst}. Looking at the failure cases for MD-GAN S2 and AE-ConvLSTM, we observe that they struggle with situations where clouds are primarily scattered. This is unsurprising given that these models tend to generate predictions that are generally clustered and moderately blurry. 
The TVL1 model projects a considerable movement of clouds that is incorrect. The underlying reason could be the dissipation of clouds from the input images used in the optical flow estimation, which would violate the constant pixel intensity assumption. The Persistence model achieves poor performance in situations with substantial motion as in Figure \ref{fig:bestworst}.

\subsubsection{Autoencoder ConvLSTM (AE-ConvLSTM)}
The AE-ConvLSTM method achieves the highest accuracy on our dataset measured both temporally and spatially on all but medium clouds. For the BSS metric, we notice superior performance relative to the persistence model with a value of 0.11. This implies the application of ConvLSTM layers for cloud-labeled satellite images do capture spatiotemporal motion to some extent. On the other hand, we see in Figure \ref{fig:forecast} that predictions become increasingly blurry over time. This is in alignment with the discussion in Section \ref{convlstm_section}. 

\newcommand{\imagebestworst}[1]{\includegraphics[width=20mm, height=18mm]{images/best_worst_new/#1}}

\begin{figure}[!t]
\scriptsize   
\renewcommand{\arraystretch}{1.2}
\setlength\tabcolsep{0pt}
\centering
\begin{tabular}{
    >{\centering\arraybackslash}m{0.5cm}
    >{\centering\arraybackslash}m{2.1cm}
    >{\centering\arraybackslash}m{2.1cm}
    >{\centering\arraybackslash}m{2.1cm}
    >{\centering\arraybackslash}m{2.1cm}
    }
& MD-GAN S2  & AE-ConvLSTM   &  TVL1   &  Persistence \\
\rotatebox{90}{Worst}  &  
\imagebestworst{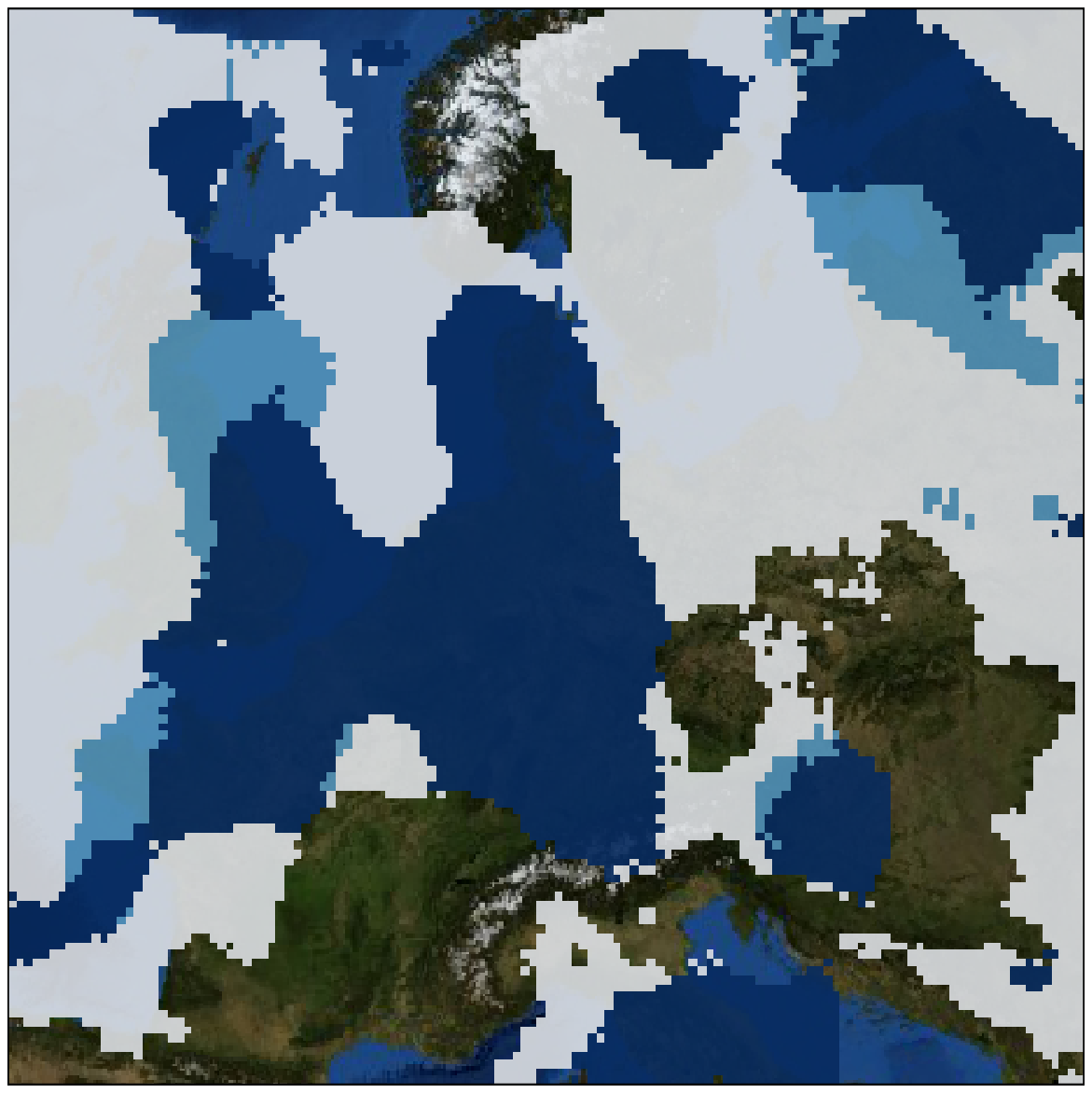}  & \imagebestworst{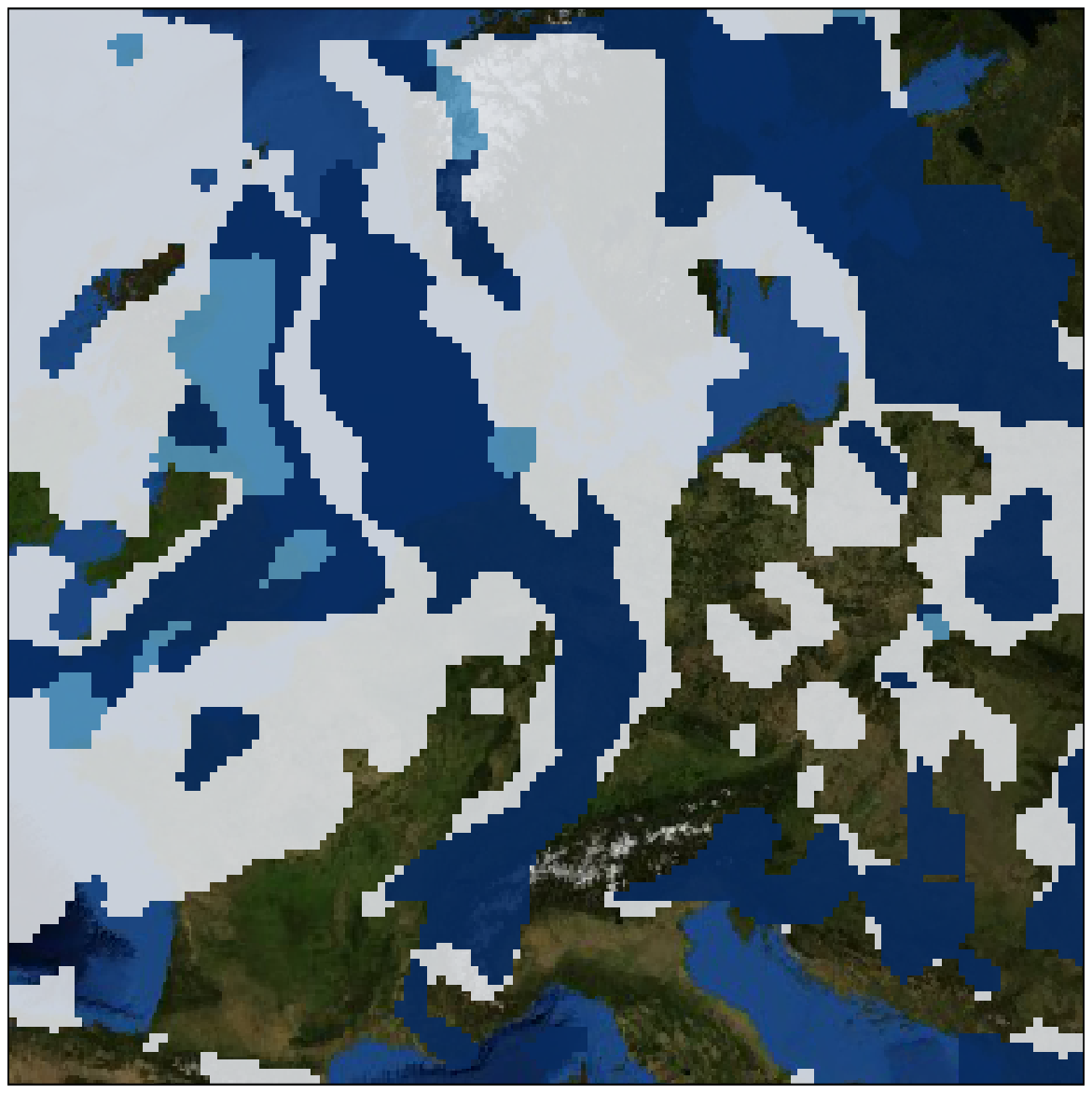}  & \imagebestworst{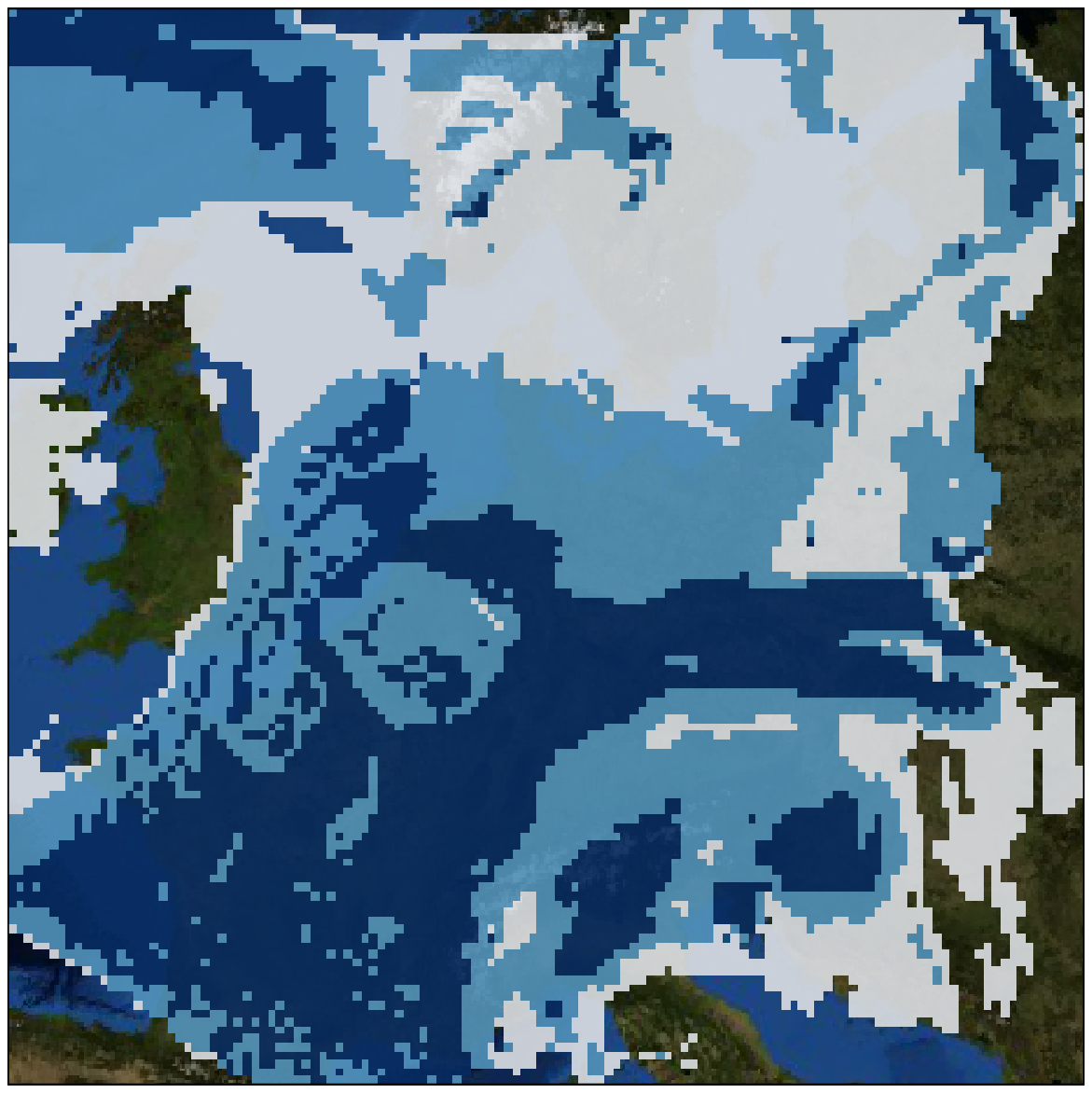} & \imagebestworst{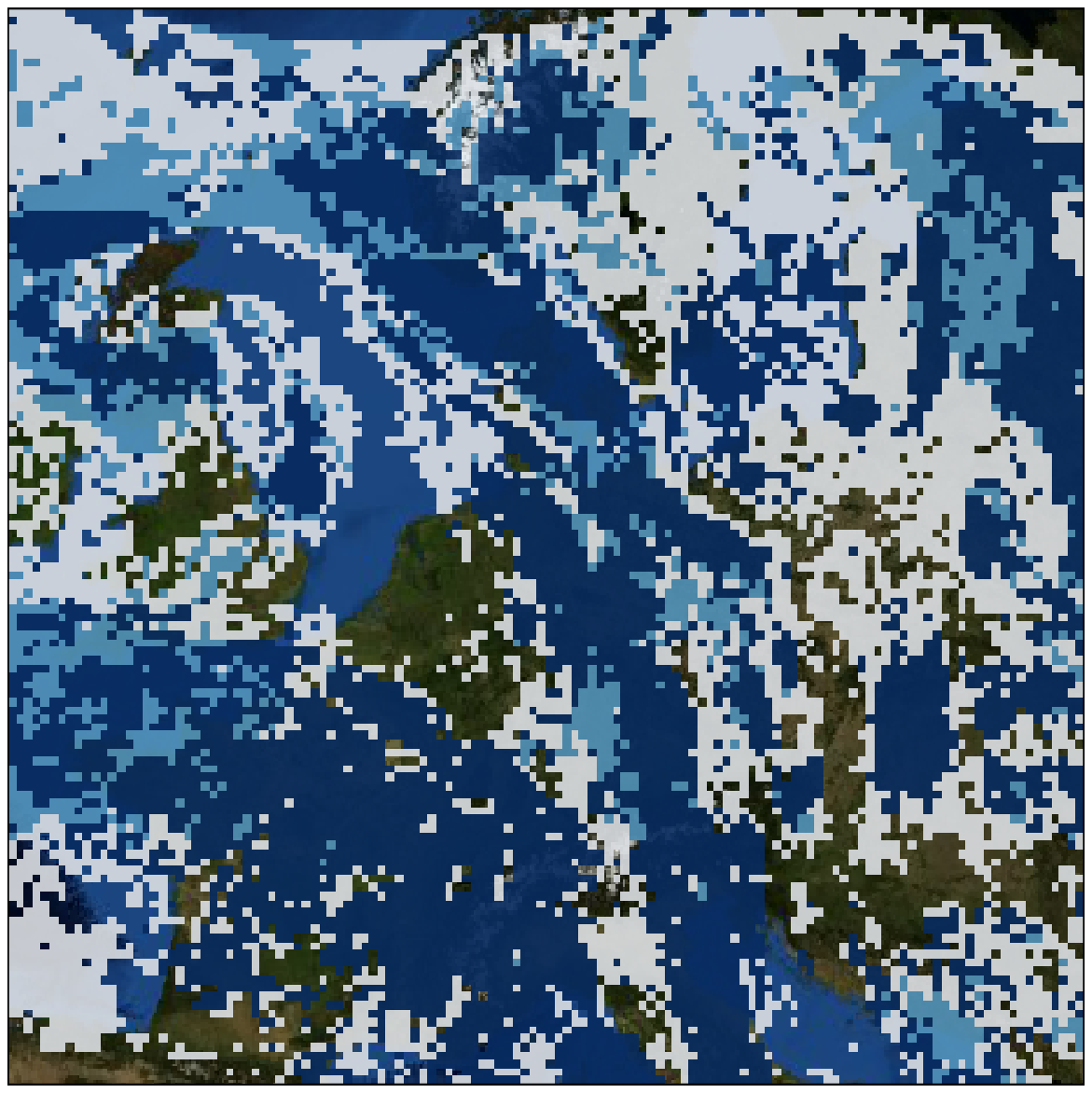} \\
\\[-1.5em]
\rotatebox{90}{Ground Truth} &  \imagebestworst{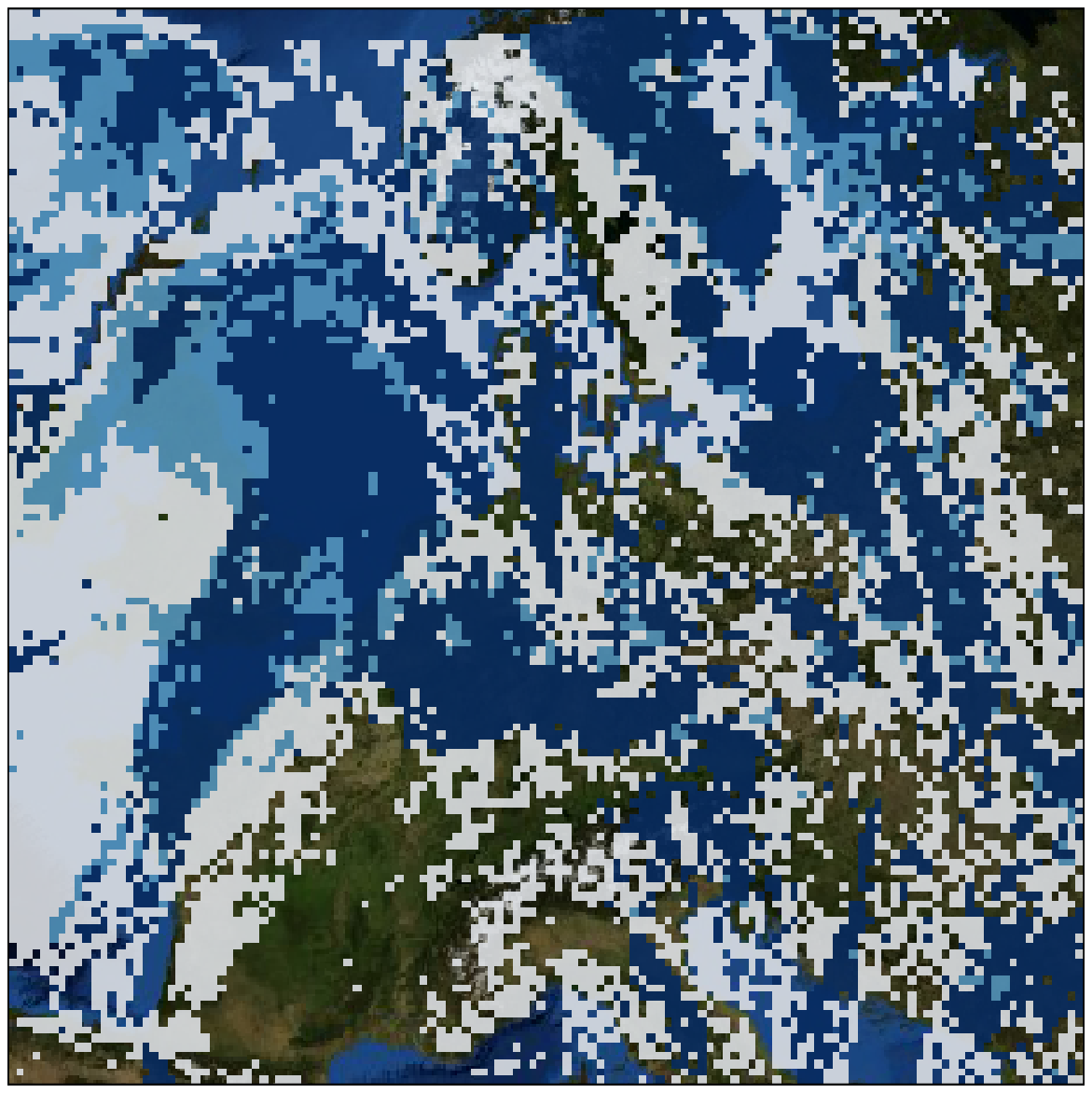}  & \imagebestworst{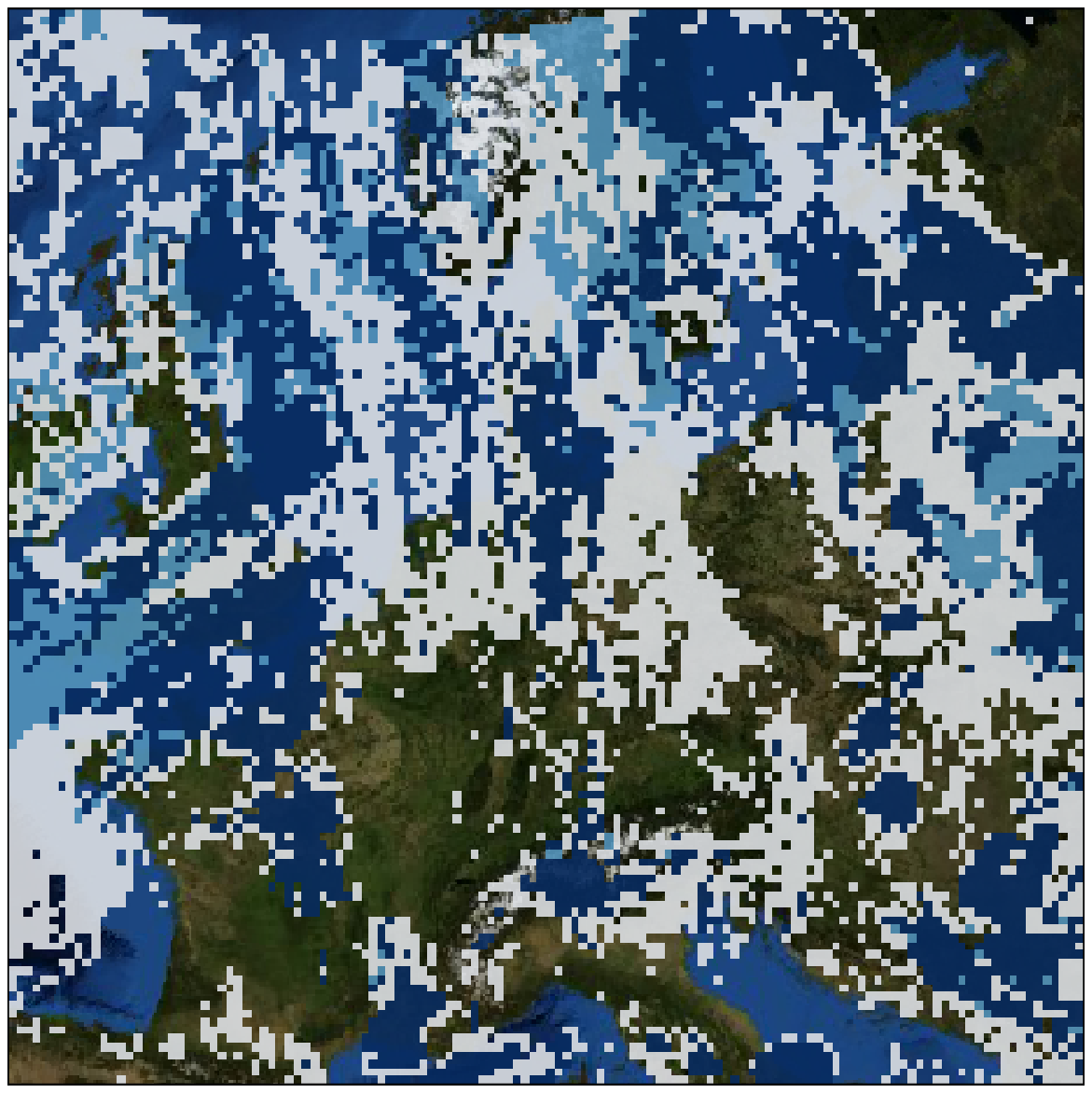}  & \imagebestworst{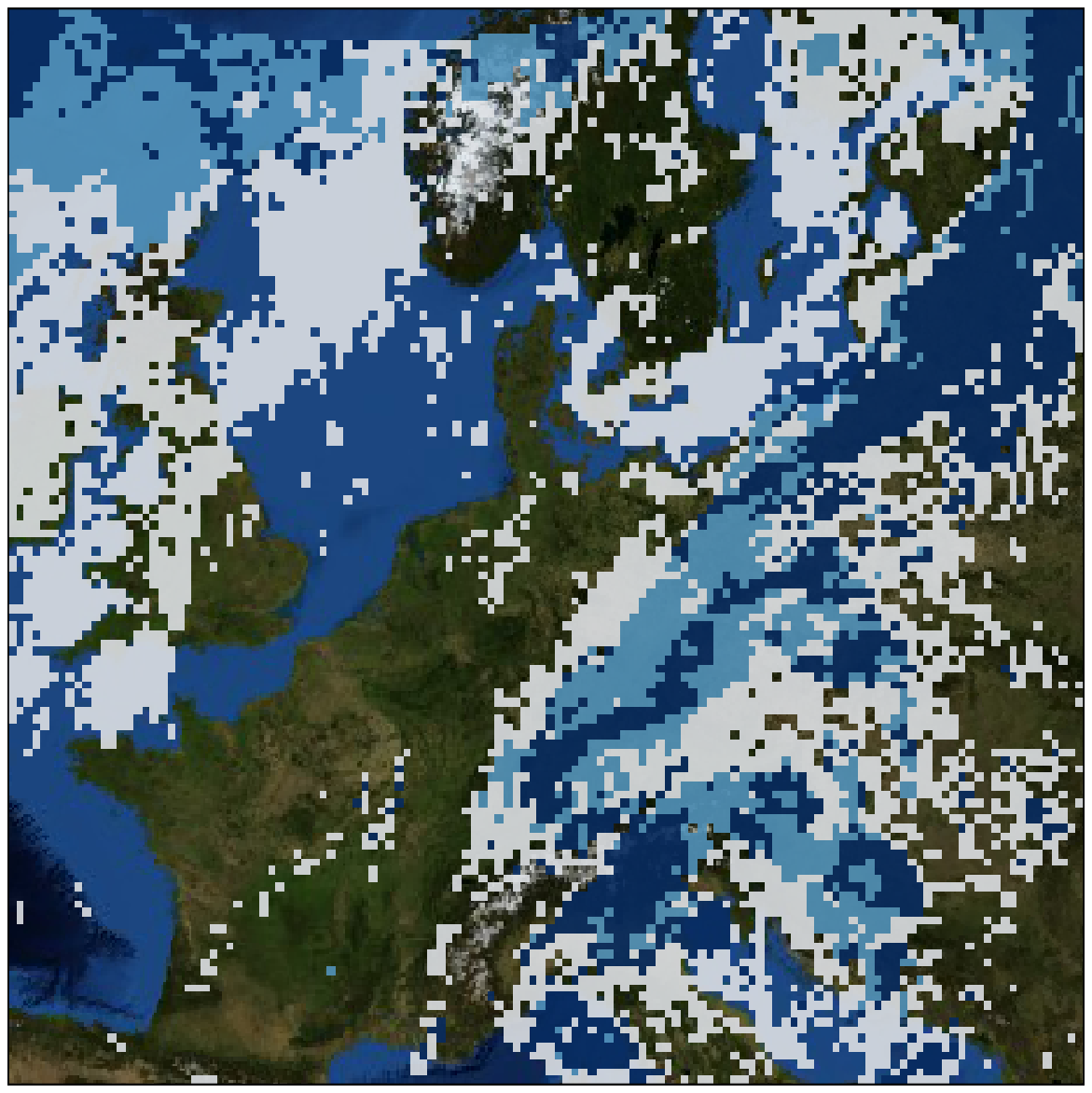} & \imagebestworst{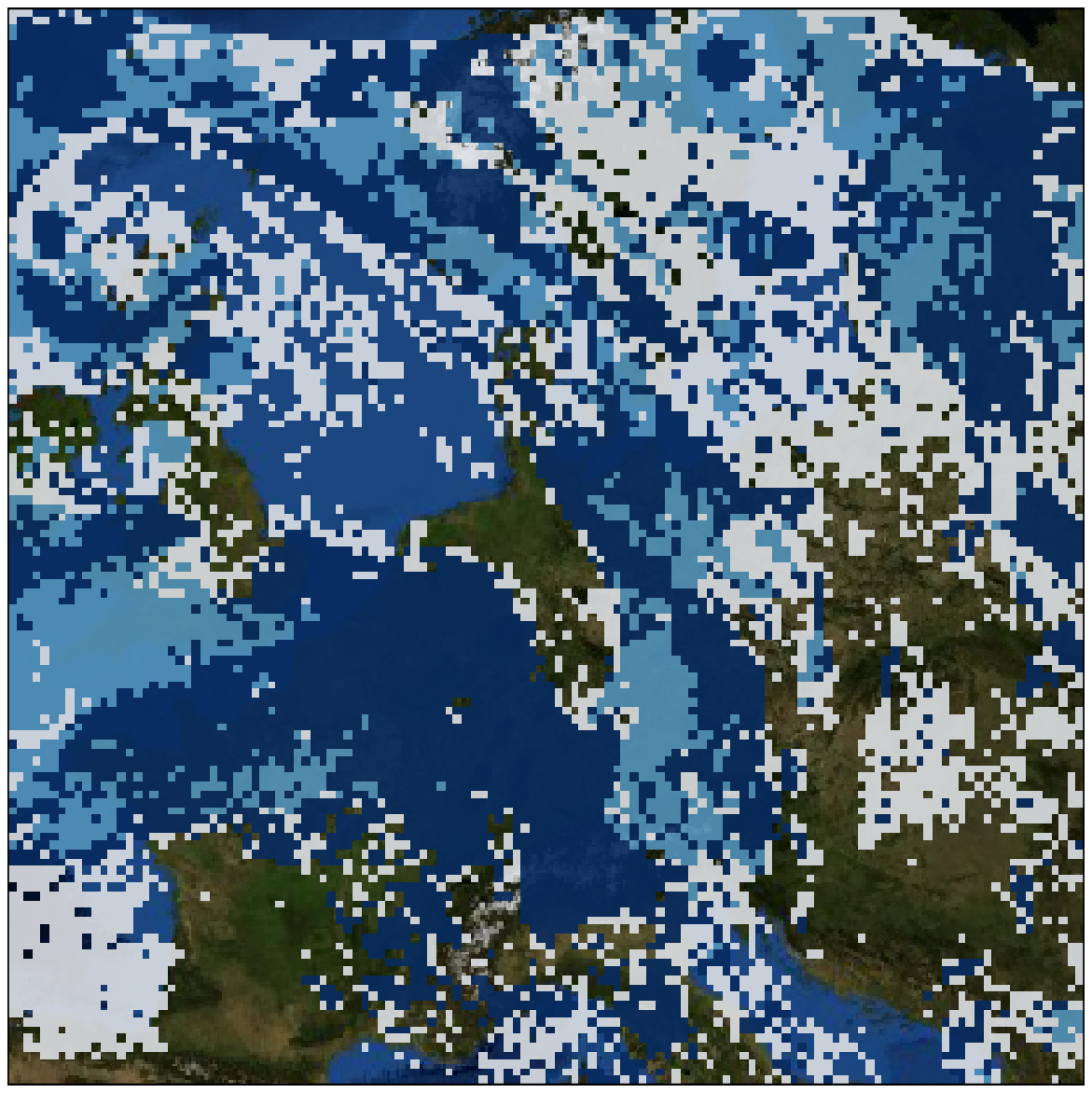} \\ 
\\[-1.5em]
\rotatebox{90}{Difference} &  \imagebestworst{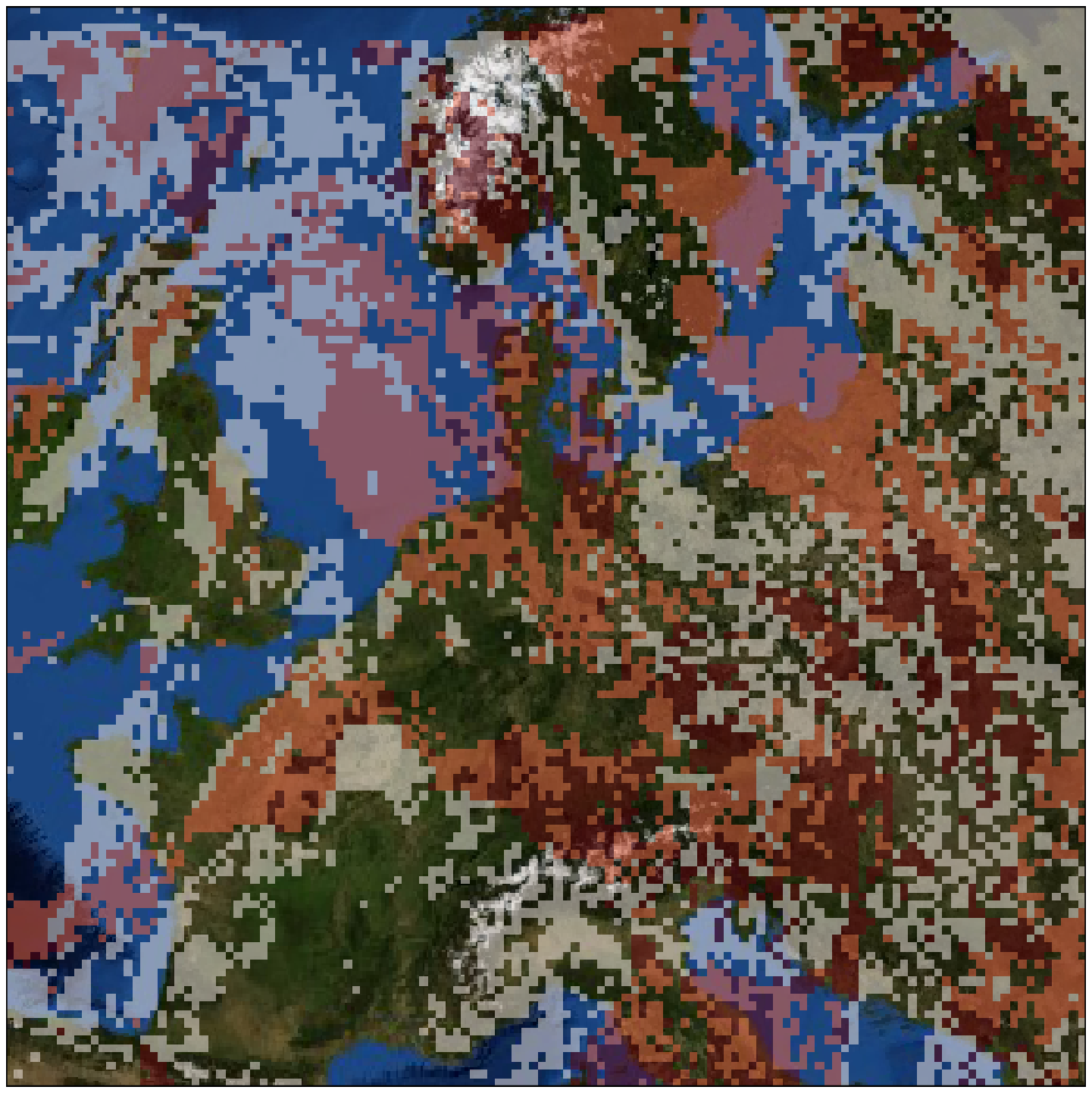}  & \imagebestworst{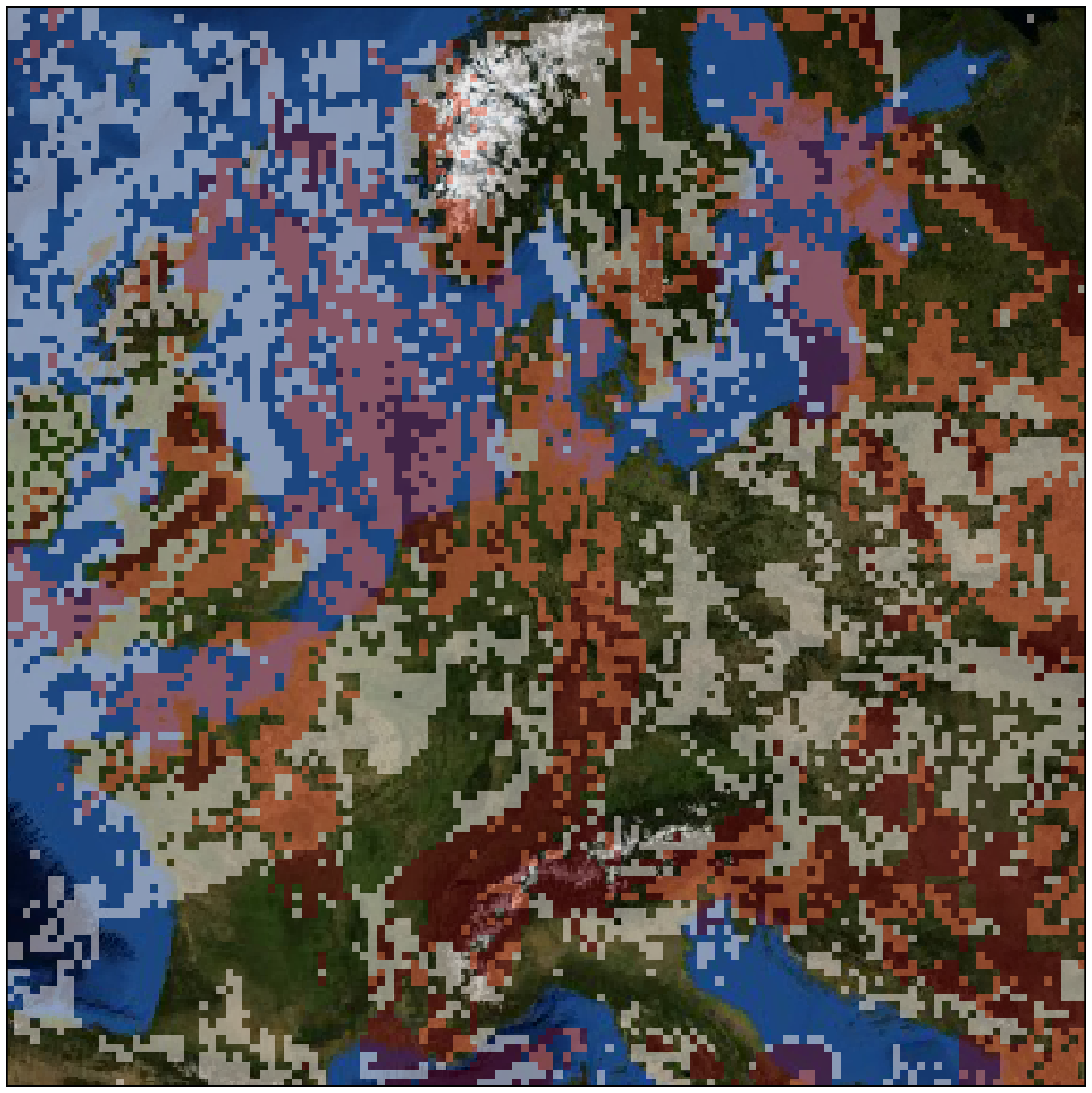}  & \imagebestworst{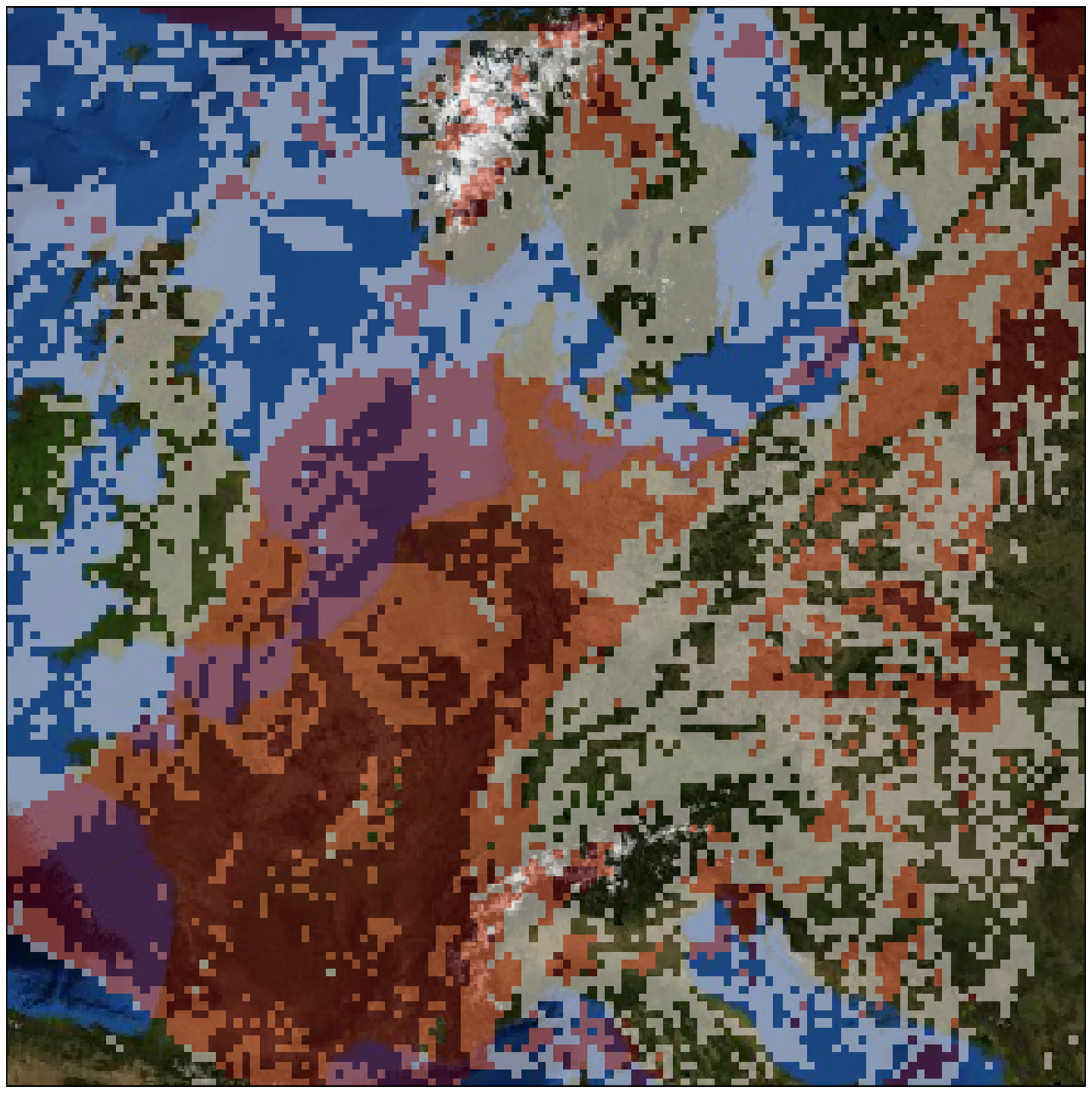} & \imagebestworst{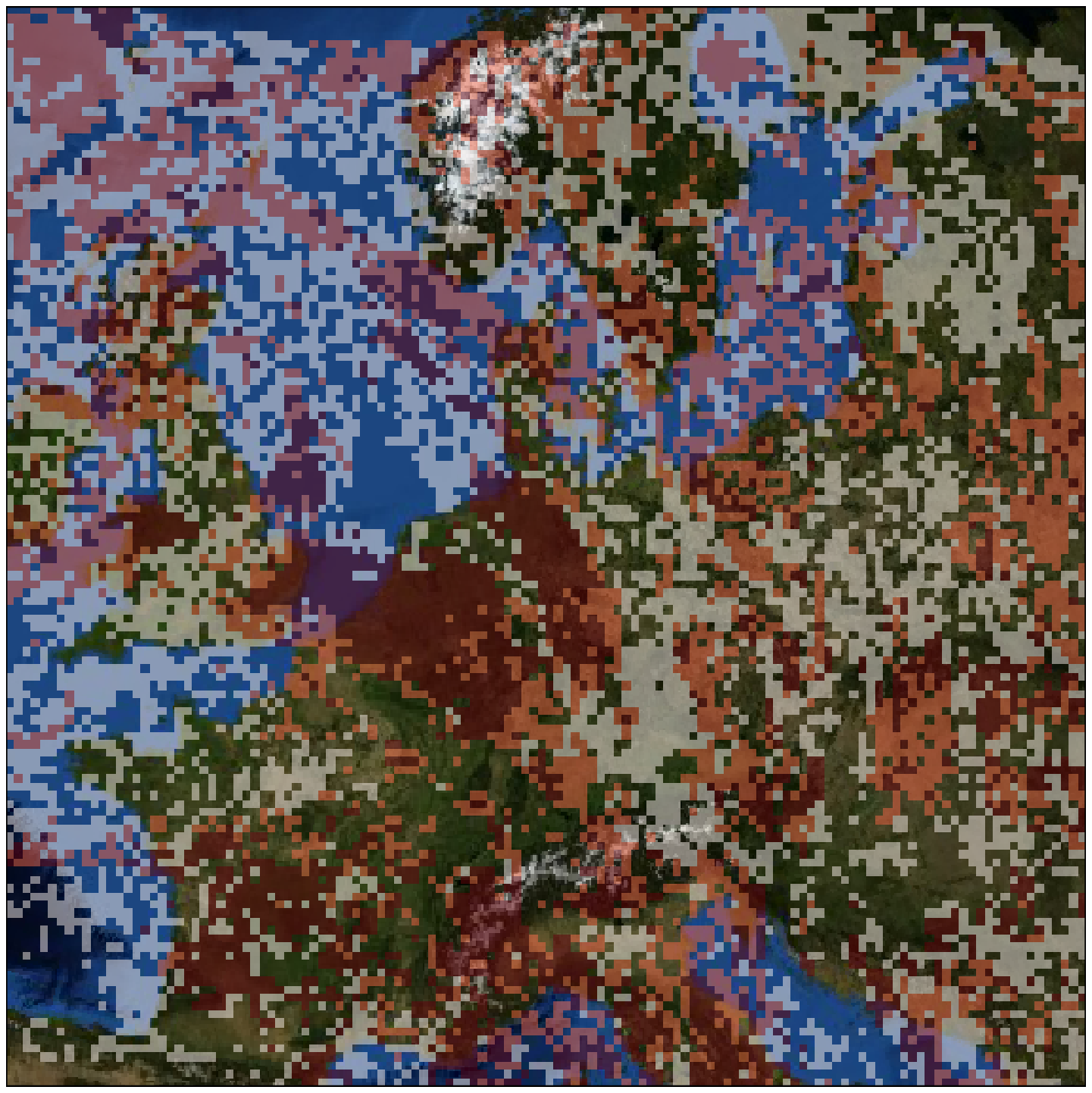} \\ 
\\[-1.5em]
\end{tabular}
\caption{Worst predictions from our test dataset on CloudCast using the proposed benchmark models compared to ground truth. Worst is defined as having the lowest mean accuracy among all test images for each model. The difference plots are calculated using the absolute difference between the predicted and ground truth images.}
\label{fig:bestworst}
\end{figure}

\subsubsection{Multi-Stage Dynamic Generative Adversarial Networks (MD-GAN)}
The MD-GAN model outperforms the persistence model with a Brier Skill Score of 0.07. The categorical accuracy is not captured well, as MD-GAN achieves the lowest accuracy for medium clouds between all our models. The temporal accuracy is closely matched to the ConvLSTM model, especially for the 2-hour forecast. Thus, by improving the stability and the initial forecasting accuracy of the MD-GAN model, we expect it could become the best and most consistent model.

\subsubsection{$\mathbf{TV\mhyphen L^{1}}$ Optical Flow (TVL1)}
The TVL1 algorithm shows marginally superior performance relative to the persistence model with a BSS of 0.02. The primary reason behind the close performance of TVL1 and Persistence relates to the choice of hyperparameters for the TVL1 algorithm, where hyperparameters yielding more static movement generally implied better performance across time. We believe the underlying reason for this result is the complexity of forecasting multi-layer clouds 16-steps ahead combined with the violation of the optical flow assumption of having constant brightness intensity over time. Compared to AE-ConvLSTM and MD-GAN, it achieves lower overall- and temporal accuracy but does reach higher accuracy for medium clouds. While optical flow methods have been popular for atmospheric forecasting, as stated in Section \ref{optflow}, their application to multi-layer cloud types has not been fully researched yet. Hence, the proposed machine learning methods currently seem more appropriate for this task given their superior performance. 

\subsubsection{Persistence}
The simple persistence model achieves relatively good results. The high short-term accuracy is not surprising given the limited cloud movement for one-hour ahead. Due to its static nature, however, it achieves the lowest accuracy near the end of our forecasting horizon.


\section{Conclusions}\label{conclusions}
We introduce a novel dataset for cloud forecasting called CloudCast, which consists of pixel-labeled satellite images with multi-layer clouds of high temporal and spatial resolution. The dataset facilitates the development and evaluation
of methods for atmospheric forecasting and video prediction in both the vertical
(height) and horizontal (latitude-longitude) domains. Four different cloud nowcasting models were evaluated on this dataset based on recent advancements in the machine learning literature for video prediction and traditional methods from the meteorology- and computer vision literature. Several evaluation metrics based on best practices in cloud forecasting studies were proposed in addition to the Peak Signal-to-Noise Ratio and Structural Similarity Index. The four models provided an initial benchmark for this dataset but showed ample room for improvement, especially for predictions near the end of our forecasting horizon. Hybrid methods combining machine-learning and NWP could be interesting approaches to address medium to long-term forecasting in a future study.

We hope this novel dataset will help advance and stimulate the development of new data-driven methods for atmospheric forecasting in a field heavily dominated by physics and numerical methods.

\ifCLASSOPTIONcaptionsoff
  \newpage
\fi


\bibliographystyle{IEEEtran}
\bibliography{IEEEabrv}

\end{document}